\documentclass{article}

\usepackage[final,main]{neurips_2026}

\usepackage[utf8]{inputenc}
\usepackage[T1]{fontenc}
\usepackage{hyperref}
\usepackage{url}
\usepackage{booktabs}
\usepackage{amsfonts}
\usepackage{amsmath}
\usepackage{amssymb}
\usepackage{amsthm}
\usepackage{mathtools}
\usepackage{microtype}
\usepackage{xcolor}
\usepackage{graphicx}
\usepackage{subcaption}
\usepackage{array}
\usepackage{makecell}
\usepackage{placeins}

\hypersetup{hidelinks, pdftitle={Signed-Permutation Coordinate Transport for RMSNorm Transformers}, pdfauthor={John Sweeney}}

\newtheorem{theorem}{Theorem}[section]
\newtheorem{proposition}[theorem]{Proposition}
\newtheorem{lemma}[theorem]{Lemma}
\newtheorem{corollary}[theorem]{Corollary}
\theoremstyle{definition}
\newtheorem{definition}[theorem]{Definition}
\newtheorem{remark}[theorem]{Remark}
\newcommand{\R}{\mathbb{R}}
\newcommand{\Sd}{S_d}
\newcommand{\Bd}{B_d}
\newcommand{\G}{\mathcal{G}}
\newcommand{\norm}[1]{\left\lVert #1 \right\rVert}
\newcommand{\inner}[2]{\left\langle #1,#2 \right\rangle}
\newcommand{\diag}{\operatorname{diag}}
\newcommand{\tr}{\operatorname{tr}}
\newcommand{\argmax}{\operatorname*{arg\,max}}
\newcommand{\argmin}{\operatorname*{arg\,min}}
\newcommand{\RMS}{\operatorname{RMS}}
\newcommand{\RMSNorm}{\operatorname{RMSNorm}}
\newcommand{\LayerNorm}{\operatorname{LayerNorm}}
\newcommand{\CKA}{\operatorname{CKA}}

\title{Signed-Permutation Coordinate Transport for RMSNorm Transformers}

\author{%
  John Sweeney\\
  Sideplane AI\\
  \texttt{john.sweeney@sideplane.ai}\\
  \url{https://sideplane.ai}\\
}

\begin{document}

\maketitle

\begin{abstract}
Modern LLM workflows increasingly move coordinate-indexed objects across checkpoints: steering vectors, sparse autoencoders, top-$k$ neuron sets, attribution lists, LoRA updates, optimizer states, and merge alignments. This is only well-posed after fixing the model's residual-stream gauge. We show that the native coordinate-preserving orthogonal gauge is architecture-dependent: LayerNorm residual charts have permutation gauge $\Sd$ up to a global sign flip, while RMSNorm residual charts with generic per-channel gains have signed-permutation gauge $\Bd=\Sd\ltimes\{\pm1\}^d$. Thus permutation-only alignment is symmetry-incomplete for RMSNorm models. We introduce sign-marginalized Hungarian matching and prove a sharp population failure mode: with decorrelated source coordinates, raw signed-correlation matching has a structural permutation-accuracy ceiling equal to the positive-sign fraction in the true gauge up to $O(d2^{-d})$, whereas sign-marginalized matching removes this obstruction. Omitting signs introduces expected squared error $2\norm{\Delta W}_F^2$ on learned weight updates and loses coordinate orientation on vectors and AdamW first moments, while RMSNorm gains and AdamW second moments permute without signs. Controlled reparameterization tests confirm these exact transport laws: a Qwen2.5-1.5B SST-2 LoRA adapter retains 0.944 accuracy from a 0.945 source under $\Bd$ transport versus 0.523 under $\Sd$, and multi-behavior steering preserves effects to $<10^{-4}$ while raw top-10 coordinate overlap drops to zero. Across natural checkpoints, inherited parameter slots serve as a strong default for same-base descendants (96.2\%--100\% retention across Qwen2.5 and DeepSeek-R1-Distill lineages), whereas cross-generation pairs support only sparse prompt-stable subsets on which negative fitted signs are causally decisive ($+0.185$ vs.\ $-0.010$ at $8\sigma$ on Qwen2/2.5-0.5B; $+0.189$ vs.\ $-0.021$ on Qwen2/2.5-1.5B). Along gauge-instrumented fine-tuning trajectories, composing local $\Bd$ gauges recovers 91.1\% of cross-run Qwen2.5-1.5B coordinates at 1,500 steps versus 60.3\% for direct endpoint matching while localizing rare step failures. Coordinate-preserving merge tests and gauge-sweep interpretability audits confirm the same boundary: coordinate names are reproducible only relative to an explicit $\Bd$ gauge.
\end{abstract}

\section{Introduction}
\label{sec:intro}

The question ``which neuron is this?'' is ill-posed without specifying a \emph{gauge}: a choice of internal coordinates within a function-preserving symmetry orbit (whose admissible transformations form the \emph{gauge group}, with any two checkpoints related by a \emph{relative gauge}). For example, coordinate 17 in one transformer checkpoint can be minus coordinate 203 in another while both compute the exact same input--output function: a positive intervention on the first coordinate must then become a negative intervention on the second, and copying the index unchanged---or matching the partner index while omitting its sign---alters the intervention. Whenever residual coordinates are relabeled by such a symmetry, external behavior is unchanged, but every coordinate-indexed artifact attached to the residual stream---ranked neuron or attribution sets, knowledge-neuron edits, SAE dictionaries, steering directions, LoRA updates, optimizer states, and merge alignments---moves with the relabeling. A coordinate-indexed claim should therefore pass a simple audit: exact gauges preserve behavior, move raw coordinate names, and restore agreement after mapping back.

Standard alignment tools such as Git Re-Basin and weight matching search over permutations [Ainsworth et al., 2023, Entezari et al., 2022]. That is correct for LayerNorm residual streams, but widely used open-weight LLMs such as Llama 2 and Qwen2.5 use RMSNorm [Touvron et al., 2023, Yang et al., 2024a, Zhang and Sennrich, 2019]. RMSNorm lacks mean-centering. Its residual stream therefore admits not only permutations but independent sign flips of every coordinate. The relevant discrete group is the hyperoctahedral group $\Bd=\Sd\ltimes\{\pm1\}^d$ of order $2^d d!$, not $\Sd$ of order $d!$. In a 4,096-dimensional residual stream, the omitted sign component fixes only one of $2^{4096}$ gauge cosets and forgoes 4,096 binary degrees of freedom; Corollary~\ref{cor:sign-omission} shows that misapplying permutation-only transport to a learned weight update introduces squared error of magnitude $2\norm{\Delta W}_F^2$ in expectation, the same order as the update's own energy. While subspace and similarity analyses are already gauge-invariant, our target is workflows that explicitly move or publish coordinate-indexed artifacts.

This paper is about the consequences of using the right group. We first prove that under normalization-covariant READ/WRITE transformations in the native $\gamma$-explicit parameterization, LayerNorm has discrete gauge $\Sd$ up to a global sign flip ($\Sd\times C_2$), while RMSNorm has full signed-permutation gauge $\Bd$, and that $\Bd$ is the maximal coordinate-preserving orthogonal subgroup whenever the joint tuples of RMSNorm gains across the residual stream are distinct (Section~\ref{sec:gauge}). Broader orthogonal freedom $O(d)$ returns only after absorbing $\gamma$ into adjacent weights [Ashkboos et al., 2024a,b, Theus et al., 2025, Zhang et al., 2025]; this is valuable when dense coordinate mixing is acceptable, but it changes the coordinate system and loses sparse coordinate identity (Theorem~\ref{thm:discrete-content}, Proposition~\ref{prop:sparse}).

We then recover candidate correspondences from activations (Section~\ref{sec:recovery}). The natural estimator is the activation-matching linear assignment used by earlier neuron-alignment and Git Re-Basin work [Li et al., 2016, Tatro et al., 2020, Ainsworth et al., 2023], solved by Hungarian assignment [Kuhn, 1955, Jonker and Volgenant, 1987], except that the objective must marginalize over sign: match columns by $|\inner{h_i}{h'_j}|$, then recover the sign from the matched entry. The absolute value is forced by $\Bd$ symmetry: under a negative gauge sign, the true partner coordinate is maximally anti-correlated. We prove in the decorrelated-coordinate population case (Corollary~\ref{thm:ceiling}) that when half the true signs are negative, signed-correlation maximization strictly avoids the negative-sign partners even with unlimited probes, capping accuracy near $1/2+O(d2^{-d})$. Because the sign factor enters weight matrices directly through $W\mapsto WG$ and $W\mapsto G^\top W$, weight-matching variants of rebasing over $\Sd$ without a sign variable inherit the exact same sign-blindness gap. Recorded-gauge LLM activations confirm this obstruction: signed-cost recovery stays near 50\%, whereas sign-marginalized recovery reaches 100\%.

We separate exact artifact covariance under controlled reparameterizations from empirical alignment across natural checkpoints. In Section~\ref{sec:artifacts}, we derive exact modulewise transport laws for learned updates, vectors, SAEs, and AdamW/SGDm optimizer states, and verify them under recorded $\Bd$ reparameterizations: on Qwen2.5-1.5B SST-2 LoRA transfer, signed $\Bd$ transport preserves 0.944 accuracy (99.7\% of adapter gain), whereas $\Sd$-only transport collapses to 0.523 ($-0.5\%$ gain, matching the unadapted base). In Section~\ref{sec:natural}, we test natural checkpoints where two models compute different functions and no exact $\Bd$ symmetry is guaranteed. Here checkpoint lineage governs the operating regime: across same-base descendants (Qwen2.5 Base/Math/Instruct, SmolLM2, Llama-3.2, TinyLlama, and \texttt{DeepSeek-R1-Distill-Qwen-1.5B}), training predominantly preserves inherited coordinate slots, making the identity map a strong default. Across distinct pretraining runs (Qwen2 vs.\ Qwen2.5 and TinyLlama old-3T vs.\ v1.1), full-space discrete maps fail held-out validation ($R^2<0$), but multi-bank stability filters isolate small prompt-stable subsets (0.5\%--3.1\% of coordinates) on which retaining negative fitted signs is causally decisive across $2\sigma$--$8\sigma$ interventions and agrees with single-coordinate-truncated $O(d)$ alignment on 10/10 negative pairs.

In Section~\ref{sec:transport}, we study parallel transport along saved checkpoint trajectories by composing local $\Bd$ matches $G_{t_k\to t_{k+1}}$. On the gauge-instrumented Qwen2.5-1.5B benchmark (1,500 steps, 18 run pairs), composing local $\Bd$ maps recovers 91.1\% of cross-run coordinates (95.3\% within-run from 6,728/6,750 exact single-step fits) versus 60.3\% for direct endpoint matching and 11.4\% for endpoint-via-base, whereas retaining the known step-0 map achieves 100\%. On Llama-3.2-1B, local composition leads endpoint matching by $+17.8$ points at 200 steps but reverses to $-3.4$ points at 1,500 steps due to localizable step failures, alongside explicit checkpoint and activation cost trade-offs. The same sign requirement governs coordinate-preserving model merging (Section~\ref{sec:merging}) and gauge-sweep audits for coordinate-indexed interpretability claims (Section~\ref{sec:audits}).

\paragraph{Relation to prior work.}
Classical results established permutation and sign symmetries in feedforward networks: Hecht-Nielsen [1990] characterized hidden-unit weight-space permutations, and Sussmann [1992] proved parameter identifiability up to hidden-node permutations and sign flips for irreducible single-hidden-layer $\tanh$ networks. In deep networks, permutation symmetries motivate mode-connectivity and rebasing methods over $\Sd$ via activation correlation or weight matching [Li et al., 2016, Garipov et al., 2018, Draxler et al., 2018, Brea et al., 2019, Tatro et al., 2020, Entezari et al., 2022, Ainsworth et al., 2023]. Soft and optimal-transport fusion relax permutations to many-to-many couplings [Singh and Jaggi, 2020, Imfeld et al., 2024, Stoica et al., 2024], while dense orthogonal methods absorb normalization gains into adjacent weights to rotate coordinates over $O(d)$ [Ashkboos et al., 2024a,b, Theus et al., 2025, Zhang et al., 2025]. Invariant similarity metrics [Raghu et al., 2017, Morcos et al., 2018, Kornblith et al., 2019] and subspace analyses [Elhage et al., 2023] are orthogonal-invariant by construction. Our contribution is the architecture-specific characterization of the native coordinate-preserving residual group ($\Bd$ for RMSNorm vs.\ $\Sd\times C_2$ for LayerNorm), the exact READ/WRITE/state covariance rules it imposes on coordinate-indexed artifacts, and the empirical boundary between exact reparameterization transport and natural-checkpoint correspondences.

\section{Native residual-stream gauge}
\label{sec:gauge}

What function-preserving coordinate transformations leave a transformer's native parameterization intact? We use row-vector notation: a residual activation matrix $H\in\R^{n\times d}$ is gauged by right multiplication $H\mapsto HG$. A signed permutation $G\in\Bd$ has exactly one nonzero entry, $\pm1$, in each row and column: it sends each coordinate axis to a single coordinate axis and preserves support size. A permutation $P\in\Sd$ is the special case with all signs $+1$. In this paper the primary group of interest is the global discrete signed-permutation action $\Bd$ on residual coordinates.

\begin{theorem}[Maximal native discrete gauge]
\label{thm:native-gauge}
Consider a transformer with residual dimension $d$, tokenwise LayerNorm or RMSNorm, standard affine projections $Y=XW^\top+b$, and attention/MLP sublayers that interact with residual coordinates through linear maps, elementwise nonlinearities, and normalization. Under the common residual-coordinate change $H\mapsto HG$ with modulewise READ/WRITE weight transforms and normalization covariance $N^G(hG)=N(h)G$, the architecture-wide signed-permutation gauge groups are $\G=\{\pm P:P\in \Sd\}\cong \Sd\times C_2$ for LayerNorm and $\G=\Bd$ for RMSNorm. For LayerNorm, the $C_2$ factor is a single global sign flip with $\beta\mapsto-\beta P$ and WRITE biases negated, so the per-index relabeling content is $\Sd$. For RMSNorm, gains permute without signs ($\gamma\mapsto\gamma[\pi^{-1}]$). Moreover, whenever the joint tuples of gain values $(\gamma_i^{(\ell)})_\ell$ across all RMSNorm layers sharing the residual stream distinguish every pair of coordinates $i\ne j$ (for which a single RMSNorm layer with distinct entries suffices), $\Bd$ is the maximal orthogonal subgroup of $O(d)$ satisfying explicit-gain normalization covariance without absorbing $\gamma$ into adjacent weights. Jointly equal gain tuples across all shared RMSNorms permit block-orthogonal rotations within their common eigenspaces, which do not preserve individual coordinate axes.
\end{theorem}

\paragraph{Proof sketch.}
Linear maps admit any orthogonal residual change $G\in O(d)$ under READ/WRITE weight transforms. A READ map consumes residual coordinates via $W^G=WG$ and $b^G=b$, giving $HG(W^G)^\top+b^G=HW^\top+b$; a WRITE map produces residual coordinates via $W^G=G^\top W$ and $b^G=bG$, giving $(XW^\top+b)G$. Consequently, projected queries, keys, values, and MLP hidden activations are invariant, and only output projections write gauged contributions back to the residual stream. Normalization restricts $G$. RMSNorm's root-mean-square denominator is $O(d)$-invariant, but preserving explicit diagonal gains requires $G^\top\diag(\gamma^{(\ell)})G$ to remain diagonal for every RMSNorm layer $\ell$ sharing the residual chart. Simultaneous diagonalization restricts each column of $G$ to a common eigenspace of $\{\diag(\gamma^{(\ell)})\}_\ell$; when the joint gain tuples are distinct across coordinates, the orthogonal solutions are precisely the signed permutations $\Bd$. LayerNorm additionally subtracts the coordinate mean $\mu(h)$, which excludes nonconstant coordinate signs and leaves only $\{\pm P:P\in\Sd\}$ (Lemma~\ref{lem:layernorm}). Full module rules, tied embeddings, and layerwise induction are proved in Appendices~\ref{app:proof-gauge}--\ref{app:rules}.

The native-vs.-absorbed distinction defines the boundary with dense orthogonal methods. SliceGPT and QuaRot use $\RMSNorm(XQ)Q^\top=\RMSNorm(X)$ after absorbing diagonal gains into neighboring weight matrices [Ashkboos et al., 2024a,b]; generalized linear mode connectivity and rotation-based fusion use related $O(d)$ transformations [Theus et al., 2025, Zhang et al., 2025]. When a dense rotated basis is acceptable, this gain absorption unlocks full $O(d)$ freedom, whereas transporting a sparse top-$k$ coordinate set, an index-level attribution list, or a diagonal AdamW second-moment state in native data structures requires $\Bd$ (Proposition~\ref{prop:sparse} and Remark~\ref{rem:adamw}). Note that Theorem~\ref{thm:native-gauge} characterizes normalization-covariant READ/WRITE parameter symmetries in the native chart rather than unconstrained input--output identifiability. An exact census of stored RMSNorm gain tensors across five public checkpoints (Llama-3.2-1B Base/Instruct, SmolLM2-360M Base/Instruct, and TinyLlama-1.1B; Table~\ref{tab:gain-census}, Appendix~\ref{app:gain-census}) confirms that while individual layers have floating-point ties (up to 345 coordinates), \emph{zero} coordinate pairs are tied across all $2L+1$ residual normalizations in any checkpoint, reducing the joint stabilizer strictly to $\Bd$.

\begin{theorem}[Discrete content of an orthogonal map]
\label{thm:discrete-content}
For any $Q\in O(d)$, define its nearest signed-permutation projections by $P^\star \in \argmin_{P\in\Bd}\norm{Q-P}_F$. A minimizer is obtained by sign-marginalized linear assignment on $|Q|$: choose
\[
\sigma^\star \in \argmax_{\sigma\in\Sd}\sum_{j=1}^d |Q_{\sigma(j),j}|,\qquad P^\star_{\sigma^\star(j),j}=\operatorname{sign}(Q_{\sigma^\star(j),j})
\]
(choosing either sign if that entry is zero; tied assignments yield multiple minimizers). The residual factor $R=Q(P^\star)^{-1}\in O(d)$ captures continuous coordinate mixing, with $\norm{R-I}_F=\norm{Q-P^\star}_F$, and the discrete-content ratio $\rho(Q)=d^{-1}\tr(Q^\top P^\star)\in[0,1]$ equals $1$ if and only if $Q\in\Bd$.
\end{theorem}

Theorem~\ref{thm:discrete-content} separates the two alignment regimes: the $\Bd$ factor $P^\star$ is a discrete coordinate-preserving scaffold, whereas $R=Q(P^\star)^{-1}$ is a dense rotation that mixes coordinates (proof in Appendix~\ref{app:decomp}). Numerically, applying random admissible relabelings in \texttt{float32} preserves held-out logits up to floating-point accumulation ($\le 1.28\times10^{-4}$) across all seven tested architectures---Qwen2.5-7B ($1.01\times10^{-4}$), Llama-3.1-8B ($7.44\times10^{-5}$), Llama-2-7B-chat ($1.28\times10^{-4}$), TinyLlama-1.1B ($8.58\times10^{-5}$), and T5-small ($7.63\times10^{-5}$) under $\Bd$, plus BERT-base ($2.29\times10^{-5}$) and ViT-B/16 ($4.29\times10^{-6}$) under $\Sd$---whereas off-group controls (independent signs on LayerNorm GPT-2 at $105.8$, or untransformed WRITE biases on TinyLlama at $10.1$) fail by six orders of magnitude (Figure~\ref{fig:symmetry-boundary} and Table~\ref{tab:gauge-validation-full} in Appendices~\ref{app:figures} and~\ref{app:arditi}).

\section{Sign-marginalized gauge recovery}
\label{sec:recovery}

Given paired probe activations $H_s,H_t\in\R^{n\times d}$, how can an unknown signed-permutation correspondence be recovered? We seek a recovery matrix $G\in\Bd$ with $H_tG\approx H_s$. Center columns and set $C=H_s^\top H_t$ (unnormalized centered columns maximize covariance; $\ell_2$-normalized columns maximize correlation, and we state preprocessing per experiment). For $\pi\in\Sd$ and signs $s_i\in\{\pm1\}$, minimizing $\sum_{i=1}^d \norm{H_s[:,i]-s_iH_t[:,\pi(i)]}_2^2$ is equivalent to maximizing $\sum_i s_i C_{i,\pi(i)}$. For fixed $\pi$, the optimal sign is $s_i=\operatorname{sign}(C_{i,\pi(i)})$ (either sign at zero), marginalizing out the $2^d$ sign variables and reducing joint $\Bd$ estimation to
\[
\pi^\star=\argmax_{\pi\in\Sd}\sum_{i=1}^d |C_{i,\pi(i)}|,
\qquad
s_i^\star=\operatorname{sign}(C_{i,\pi^\star(i)}),
\]
a linear assignment on $-|C|$ solved via SciPy's \texttt{linear\_sum\_assignment} [Kuhn, 1955, Jonker and Volgenant, 1987, Crouse, 2016, Virtanen et al., 2020].

\paragraph{Permutation/sign convention.}
Write $P_\pi$ for the row-action permutation matrix with $(P_\pi)_{i,\pi(i)}=1$, so $(hP_\pi)_{\pi(i)}=h_i$. Matching returns source-indexed signs $s_i=\operatorname{sign}(C_{i,\pi(i)})$ and $S_s=\diag(s_i)$. The forward gauge is $G_{\rm fwd}=S_sP_\pi$ ($H_t\approx H_sG_{\rm fwd}$) and the recovery action is $G_{\rm rec}=P_\pi^\top S_s=G_{\rm fwd}^{-1}$ ($H_tG_{\rm rec}\approx H_s$). Equivalently, with target-indexed signs $S_t=P_\pi^\top S_sP_\pi$, we have $G_{\rm fwd}=P_\pi S_t$; formulas writing $G=P\diag(s)$ use target-indexed signs, and Section~\ref{sec:transport} uses forward subscripts.

\begin{corollary}[Structural ceiling in the decorrelated-coordinate case]
\label{thm:ceiling}
Suppose two activation matrices are related by a true signed permutation $(\pi^\star,s^\star)$, where probe rows are independent, identically distributed samples with finite second moments and source coordinates are uncorrelated with positive variances $\sigma_i^2>0$. Compare the signed-cost and sign-marginalized estimators $\hat\pi_{\rm signed}=\argmax_{\pi\in\Sd}\sum_{i=1}^d C_{i,\pi(i)}$ and $\hat\pi_{\rm abs}=\argmax_{\pi\in\Sd}\sum_{i=1}^d |C_{i,\pi(i)}|$. As the probe sample size $N\to\infty$, the permutation accuracy of $\hat\pi_{\rm signed}$ converges to the positive-sign fraction $|\{i:s_i^\star=+1\}|/d$ whenever at least two true signs are negative; under independent uniform signs $s_i^\star\sim\mathrm{Unif}\{\pm1\}$, its limiting expected accuracy is $1/2+O(d2^{-d})$. By contrast, the sign-marginalized estimator $\hat\pi_{\rm abs}$ converges to $\pi^\star$ under the assignment-margin condition of Lemma~\ref{lem:assignment-margin}.
\end{corollary}

Corollary~\ref{thm:ceiling} (proved in Appendix~\ref{app:decomp}) explains why permutation-only RMSNorm alignment fails even with unlimited probes: when a coordinate's true sign is negative, its true partner is the most negative entry in its correlation row, so maximizing raw signed correlation is designed to avoid it and deranges all negative-sign coordinates whenever at least two negative signs exist. Correlated residual coordinates can shift the numerical ceiling, but not the sign-marginalization requirement.

In controlled recovery experiments on 7B/8B RMSNorm models with a recorded $\Bd$ basis change, naive Hungarian matching on signed cost recovers only 49.4\%--49.9\% of the permutation and 50.1\% of signs, matching the predicted $1/2$ obstruction, whereas sign-marginalized matching recovers 100.0\% of both on the same activations (Table~\ref{tab:signed-vs-abs} in Appendix~\ref{app:arditi}). Probe requirements depend sharply on the regime (summarized in Table~\ref{tab:probe-regimes} and Tables~\ref{tab:tiny-probe}--\ref{tab:natural-probe} in Appendix~\ref{app:probe}): known-gauge exact recovery reaches 100\% with as few as 22 probe tokens on TinyLlama-1.1B (22--500 tokens across models), local trajectory transitions use $\sim$500 tokens per edge, same-base natural descendants stabilize at 36K--99K tokens, and cross-generation natural subsets require 3--4 prompt banks with held-out validation. Once $G$ is recovered, gradient-times-activation attribution is also restored: signs cancel in $(hG)\odot((\nabla_h L)G)=(h\odot\nabla_h L)P$, so $P$ maps top-$k$ attribution indices back.

\section{Transport rules and controlled artifact tests}
\label{sec:artifacts}

How must a coordinate-indexed artifact transform when a checkpoint's residual coordinates are relabeled by $G\in\Bd$? Under an exact reparameterization, each parameter's transformation rule is fixed by its modulewise READ or WRITE role.

\begin{proposition}[Gauge-covariance of learned weight updates]
\label{prop:update-covariance}
Let $\theta$ be a transformer parameterization satisfying the assumptions of Theorem~\ref{thm:native-gauge}, and let $G\in\G$ be an admissible discrete gauge. Write $T_G$ for the modulewise READ/WRITE/RMSNorm transform of Theorem~\ref{thm:native-gauge} acting on weights. Then $T_G$ is linear: for any weight delta $\Delta\theta$ on the same parameter set, however produced, $T_G(\theta+\Delta\theta)=T_G(\theta)+T_G^{\mathrm{lin}}(\Delta\theta)$, where $T_G^{\mathrm{lin}}$ applies the same modulewise READ/WRITE rules to $\Delta\theta$ as a free linear map. In particular, for a low-rank LoRA factor $\Delta W=BA$ on a residual-stream projection, $\mathrm{READ}$ maps via $A^G=AG,\,B^G=B$, and $\mathrm{WRITE}$ maps via $A^G=A,\,B^G=G^\top B$.
\end{proposition}

Since $T_G^{\mathrm{lin}}$ is linear in $G=PS$, $P$-only transport reverses every negative-sign coordinate.

\begin{corollary}[Update error from omitting signs]
\label{cor:sign-omission}
For a READ-side update $\Delta W$ under RMSNorm gauge $G=PS$ with permutation $P$ and signs $S=\diag(s)$, $s_j\in\{\pm1\}$, the squared Frobenius update error of $\Sd$-only transport (correct permutation, signs forced to $+1$) is $\|\Delta WPS-\Delta WP\|_F^2=4\sum_{j:s_j=-1}\|(\Delta WP)_{:,j}\|_2^2$, with expectation $\mathbb{E}_S[\cdot]=2\|\Delta W\|_F^2$ over independent uniform signs ($P$ and $\Delta W$ fixed). For a WRITE-side update, $\|G^\top\Delta W-P^\top\Delta W\|_F^2=4\sum_{j:s_j=-1}\|(P^\top\Delta W)_{j,:}\|_2^2$, also with expectation $2\|\Delta W\|_F^2$.
\end{corollary}

For a single residual-stream vector, $\Sd$-only transport preserves support but loses orientation.

\begin{lemma}[Artifact orientation]
\label{lem:artifact-orientation}
Let $G=P\diag(s)\in\Bd$ with $s\in\{\pm1\}^d$. For a nonzero residual-aligned vector artifact $v\in\R^d$---a steering vector, cached activation, linear-probe direction, SAE encoder row, or LoRA READ-side row---the correct $\Bd$ transport is $vG$, while permutation-only ($\Sd$) transport is $vP$ (WRITE-side matrices and SAE decoder columns transform via $G^\top$; summary in Table~\ref{tab:transport-rules}, Appendix~\ref{app:rules}). These satisfy $\|vG-vP\|_2^2=4\sum_{j:s_j=-1}(vP)_j^2$ and $\cos\angle(vG,vP)=\sum_j s_j(vP)_j^2/\sum_j(vP)_j^2$. For independent balanced random signs, the cosine has mean zero and variance $\sum_j(vP)_j^4/\|v\|_2^4$, concentrating near zero when energy is spread across coordinates. Permutation-only transport retains mapped support but loses orientation.
\end{lemma}

\paragraph{Controlled LoRA, steering, and SAE tests.}
\label{sec:controlled-artifacts}
We test Proposition~\ref{prop:update-covariance} and Corollary~\ref{cor:sign-omission} by applying a recorded random $\Bd$ reparameterization to a base model and transferring a trained SST-2 LoRA adapter into the resulting coordinate chart (Table~\ref{tab:artifact-covariance}). On Qwen2.5-1.5B, $P$-only transport collapses adapter accuracy to 0.523, matching the unadapted target base (0.525, $-0.5\%$ gain preserved) and falling below raw unchanged transfer (0.622, $23.0\%$ gain preserved), whereas signed $\Bd$ transport achieves 0.944 ($99.7\%$ of adapter gain from the 0.945 source). On TinyLlama-1.1B, signed $\Bd$ transport preserves $100.0\%$ of adapter gain (0.950 vs.\ 0.740 base), compared with $26.2\%$ (0.795) for $P$-only and $14.3\%$ (0.770) for raw transfer. All arms evaluate the same trained adapter at identical precision, isolating sign omission.

\begin{table}[t]
\centering\small
\caption{Controlled SST-2 LoRA transfer under exact signed reparameterizations $\Bd$ (Qwen2.5-1.5B: 872 examples; TinyLlama-1.1B: 200 examples; higher is better), with percentage of source adapter gain preserved over the unadapted target base ($P$-only forces signs to $+1$; Signed $G$ applies $\Bd$).}
\label{tab:artifact-covariance}
\begin{tabular}{lrrrrrr}
\toprule
Model & Source & Target base & Raw (gain) & $P$-only (gain) & Signed $G$ (gain)\\
\midrule
Qwen2.5-1.5B & 0.945 & 0.525 & 0.622 (23.0\%) & 0.523 ($-0.5\%$) & 0.944 (99.7\%)\\
TinyLlama-1.1B & 0.950 & 0.740 & 0.770 (14.3\%) & 0.795 (26.2\%) & 0.950 (100.0\%)\\
\bottomrule
\end{tabular}
\end{table}

Controlled multi-behavior steering tests across sentiment, sycophancy, and refusal on TinyLlama-1.1B verify Lemma~\ref{lem:artifact-orientation}: under exact $\Bd$ transport, steering effects are preserved across all layers with mean absolute error $<3\times10^{-5}$ ($1.15\times10^{-5}$ sentiment, $2.85\times10^{-5}$ sycophancy, $1.98\times10^{-5}$ refusal; Table~\ref{tab:steering-behaviors}, Appendix~\ref{app:steering}), while raw top-10 coordinate overlap drops to $0.00/10$--$0.05/10$. Transporting an SAE's encoder ($W_{\rm enc}\mapsto W_{\rm enc}G$) and decoder ($W_{\rm dec}\mapsto G^\top W_{\rm dec}$, $b_{\rm dec}\mapsto b_{\rm dec}G$) likewise preserves feature activations $z^G(hG)=z(h)$ and reconstruction loss (Proposition~\ref{prop:sae-invariance}, Appendix~\ref{app:sae}).

\begin{remark}[Stateful resumption: $\Bd$ is the maximal gauge for native AdamW]
\label{rem:adamw}
\label{sec:stateful}
Under an exact $\Bd$ reparameterization $G=P\diag(s)$, gradients transform with their parameters, so AdamW's first moment $m_W$ inherits the modulewise READ/WRITE/NORM rules $m_W\mapsto T_G(m_W)$. Because squaring a signed-permuted gradient permutes its squares, $(gG)^{\odot2}=g^{\odot2}|G|$, the second moment $v_W$ is sign-blind and only permutes: $v_W\mapsto T_{|G|}(v_W)=T_P(v_W)$. Elementwise square roots, division, bias correction, scalar weight decay, and global norm clipping commute with this action in exact arithmetic (up to floating-point kernel and stochastic-rounding discrepancies; we state this as an algebraic characterization without a resumed-training experiment). Moreover, $\Bd$ is the maximal orthogonal subgroup preserving arbitrary diagonal second-moment states: a dense rotation $Q\notin\Bd$ introduces off-diagonal entries $Q^\top\diag(v)Q$ not stored by diagonal AdamW.
\end{remark}

\section{Natural checkpoints: inherited slots, validated subsets, and failure}
\label{sec:natural}

Does a fitted signed-permutation map help when the target checkpoint was produced by natural training rather than an explicit reparameterization? Because two naturally trained checkpoints compute different functions, Theorem~\ref{thm:native-gauge} does not guarantee that their activations are related by a global signed permutation. The answer depends on checkpoint ancestry and on what held-out probes and interventions validate.

\paragraph{Same-base descendants and checkpoint phylogeny.}
Across same-base continued-pretraining and instruction-tuned descendants---Qwen2.5-1.5B Base/Math and Base/Instruct, Qwen2.5-0.5B Base/Instruct, SmolLM2-360M Base/Instruct, Llama-3.2-1B Base/Instruct, and TinyLlama v1.1 versus its Math/Code and Chinese continued-pretraining variants---sign-marginalized activation matching predominantly retains the inherited coordinate indices with $+1$ signs (96.2\% diagonal retention on Qwen2.5-1.5B Base/Math; 100\% on the Base/Instruct and TinyLlama continued-training pairs; Appendix~\ref{app:natural}). This slot inheritance also tracks multi-stage model phylogeny: \texttt{DeepSeek-R1-Distill-Qwen-1.5B} descends from \texttt{Qwen2.5-Math-1.5B} and inherits its parameter slots, and same-index diagonal matching identifies that true parent with a 1.000 self-match fraction, versus 0.915 for the nearest sibling (\texttt{Qwen2.5-Coder-1.5B}) and 0.002 for the cross-generation \texttt{Qwen2-1.5B} checkpoint. In downstream artifact transfers between Qwen2.5-1.5B Base and Math (Table~\ref{tab:natural-transfer}), retaining the inherited coordinate slots (Identity) matches refitted $\Sd$ within one evaluation example on Math$\to$Base LoRA transfer (0.844 vs.\ 0.845; unadapted Base 0.549) and leads on Base$\to$Math LoRA (0.654) and SAE reconstruction (0.673 NMSE vs.\ 0.817--1.115), while 0/29 single-bank negative signs in the fitted $\Bd$ map survive across all four prompt banks (Appendix~\ref{app:natural}).

\begin{table}[t]
\centering\small
\caption{Natural Qwen2.5-1.5B Base/Math artifact transfers. Identity retains inherited slots; $\Sd$, $\Bd$, and $O(d)$ are fitted on WikiText-103 activations. LoRA SST-2 accuracy and steering effect (log-prob margin shift) are higher-is-better; SAE NMSE is lower-is-better.}
\label{tab:natural-transfer}
\begin{tabular}{lrrrr}
\toprule
Artifact and transfer direction & Identity & Fitted $\Sd$ & Fitted $\Bd$ & Dense $O(d)$\\
\midrule
LoRA accuracy, Math $\to$ Base (unadapted: 0.549) & 0.844 & 0.845 & 0.842 & 0.833\\
LoRA accuracy, Base $\to$ Math (unadapted: 0.630) & 0.654 & 0.650 & 0.649 & 0.653\\
SAE reconstruction NMSE, Base $\to$ Math & 0.673 & 0.817 & 0.942 & 1.115\\
Sentiment steering effect, Base $\to$ Math & 1.176 & 1.133 & 1.144 & 1.342\\
Sentiment steering effect, Math $\to$ Base & 0.938 & 0.891 & 0.851 & 0.794\\
\bottomrule
\end{tabular}
\end{table}

\paragraph{Prompt-stable subsets and causal sign validation.}
Between distinct pretraining runs (Qwen2 vs.\ Qwen2.5 and TinyLlama old-3T vs.\ v1.1), no global permutation holds, so we filter for coordinates that select the exact same partner and sign across three distinct prompt domains (WikiText-103 W1, GSM8K, and MBPP, after dropping the first two token positions and regressing out activation norm), reserving a disjoint WikiText split (W2) strictly for held-out causal evaluation. On Qwen2-0.5B versus Qwen2.5-0.5B (pooled across layers 6, 12, and 18), this filter retains 28 of 896 coordinate pairs (3.1\% coverage), of which 10 have negative fitted signs (18 of the 28, including 7 of the 10 negative pairs, also select the identical partner in single-layer fits at every depth). On TinyLlama-1.1B old-3T (\texttt{step-1431k-3T}) versus \texttt{TinyLlama\_v1.1} (layers 5, 11, and 16), the same protocol retains 10 of 2,048 pairs, 3 with negative signs. On Qwen2-1.5B versus Qwen2.5-1.5B (layers 7, 14, and 21), four-bank agreement selects 15 of 1,536 pairs (6 negative); refitting on the three non-evaluation banks returns 17 pairs containing all 15.

To test whether these negative fitted signs reflect true opposite coordinate orientations rather than fitting noise, we inject a single-coordinate activation intervention of dose $a\sigma$ ($a\in\{2,4,8\}$ standard deviations) at the source coordinate and measure the correlation of the resulting output-logit shift on held-out W2 tokens when the matched target coordinate is perturbed with the fitted sign ($\Bd$), with the sign removed ($+a\sigma$ at the same partner), under a separately fitted $\Sd$ matcher, or at a random coordinate (Table~\ref{tab:natural-signs} in Appendix~\ref{app:natural}). Across all three model pairs, signed $\Bd$ perturbation beats sign removal on every negatively oriented pair. On Qwen2/2.5-0.5B (10/10 pairs at every dose), mean logit-shift correlation is $+0.132$ versus $-0.112$ at $2\sigma$, $+0.154$ versus $-0.083$ at $4\sigma$, and $+0.185$ versus $-0.010$ at $8\sigma$ (with fitted $\Sd$ at $+0.047$ and random at $+0.022$). On TinyLlama old-3T/v1.1 (3/3 pairs at both doses), signed $\Bd$ reaches $+0.118$ versus $-0.106$ ($\Sd$ $+0.061$, random $-0.016$) at $2\sigma$ and $+0.169$ versus $-0.051$ ($\Sd$ $+0.097$, random $+0.002$) at $8\sigma$; on Qwen2/2.5-1.5B at $8\sigma$ (6/6 negative pairs, 15/15 overall), signed $\Bd$ reaches $+0.189$ versus $-0.021$ for sign removal (shuffled control $+0.055$, random coordinate $-0.011$; $+0.140$ vs.\ $-0.004$ across all 15 pairs).

\paragraph{Support-matched comparison with dense $O(d)$ alignment.}
When unconstrained dense mixing is allowed, 3-layer pooled $O(d)$ Procrustes alignment achieves slightly higher mean intervention correlation than one-coordinate $\Bd$ transport at $8\sigma$ on Qwen2/2.5-0.5B ($+0.209$ vs.\ $+0.185$ on the 10 negative pairs; under single-layer Procrustes, $+0.175$ vs.\ $+0.185$ on the 10 negative pairs and $+0.193$ vs.\ $+0.179$ across all 28 pairs), because it spreads a single source coordinate across all $d=896$ target coordinates. When we restrict the fitted single-layer $O(d)$ map to a matching support budget of $k=1$ target coordinate (Theorem~\ref{thm:discrete-content}), it selects the \emph{exact same partner coordinate and negative sign} as $\Bd$ on all 10/10 sign-informative Qwen2/2.5-0.5B pairs (both scoring $+0.185$) and on 24/28 stable pairs overall ($+0.163$ vs.\ $+0.179$ for $\Bd$), requiring $k=4$ target coordinates ($+0.180$ all, vs.\ $+0.193$ at full $k=896$) to match the one-coordinate map.

\paragraph{Global map rejection and operational scope.}
Outside these prompt-stable subsets, full-space discrete maps between independently pretrained checkpoints do not generalize: on Qwen2-1.5B versus Qwen2.5-1.5B, a full-space map fit on 70,656 pooled activation rows yields held-out $R^2=-0.971$ for Identity, $-0.742$ for $\Sd$, and $-0.722$ for $\Bd$ (whereas dense $O(d)$ and regularized ridge regression reach $0.146$ and $0.467$ by mixing coordinates). Moreover, a $\sim 50\%$ negative-sign fraction in a raw Hungarian fit is never sufficient evidence of natural sign flips: permuting token rows to destroy pairing produces $0.497\pm 0.008$ negative signs by chance (Appendix~\ref{app:natural}). In practice, use exact $\Bd$ transport when a reparameterization is recorded, retain inherited slots for same-base descendants, and restrict cross-generation discrete claims to prompt-stable, causally validated pairs.

\section{Parallel transport along checkpoint trajectories}
\label{sec:transport}
\label{sec:trajectory}

When two checkpoints lie along a training trajectory, can composing local step-to-step maps recover correspondences that direct endpoint matching misses? Endpoint matching estimates a single correspondence between two checkpoints; trajectory transport composes local estimates along saved intermediate checkpoints. For a trajectory $t_0,t_1,\ldots,t_T$, estimate each local step $G_{t_k\to t_{k+1}}\in\Bd$ by sign-marginalized matching with $H_{t_k}G_{t_k\to t_{k+1}}\approx H_{t_{k+1}}$. Within this section we use forward-direction subscripts $G_{a\to b}$ with $H_aG_{a\to b}\approx H_b$ (the recovery action of Section~\ref{sec:recovery} is its transpose). For a single trajectory and for two runs $a,b$ initialized from a shared base checkpoint, set
\[
G_{0\to T}=G_{t_0\to t_1}G_{t_1\to t_2}\cdots G_{t_{T-1}\to t_T},
\qquad
\hat G_{a\to b}=\left(G^{(a)}_{0\to T}\right)^{-1}G^{(b)}_{0\to T}.
\]
Function-level equivalence alone cannot define canonical transport between independent endpoints.

\begin{proposition}[No canonical function-level transport between independent endpoints]
\label{prop:no-canonical}
Let $\G$ be a nontrivial gauge group acting on parameter space $\Theta$ with $\theta\sim\theta'$ iff $f(\cdot;\theta)=f(\cdot;\theta')$. No recovery map $A:\Theta\times\Theta\to\G$ can simultaneously satisfy (i) function-level factoring, $A(\theta_1,\theta_2)=A(\theta_1,\theta_2H)$ for all $H\in\G$, and (ii) gauge-equivariance, $A(\theta_1,\theta_2H)=H^{-1}A(\theta_1,\theta_2)$, since the two together force $H=I$ for every $H\in\G$. Hence no canonical coordinate transport between independently specified parameterizations exists from function-level information alone.
\end{proposition}

Shared-base trajectories bypass this obstruction by anchoring runs to a common origin and composing local transitions on a checkpoint graph, forming a consistent atlas when cycle holonomy vanishes.

\begin{proposition}[Atlas consistency iff zero holonomy]
\label{prop:holonomy}
Let $V$ be a finite set of model checkpoints and $E$ a connected set of edges, each with directed label $\smash{\hat G_{ij}}\in\Bd$ and reverse label $\smash{\hat G_{ji}=\hat G_{ij}^{-1}}$. There exist per-checkpoint gauges $R_i\in\Bd$ such that $\smash{\hat G_{ij}=R_i^{-1}R_j}$ for every edge if and only if every cycle has identity holonomy $\smash{\hat G_{i_1i_2}\cdots \hat G_{i_ki_1}}=I$.
\end{proposition}

Composing consecutive checkpoints from a shared base forms a tree on which cycle holonomy vanishes by construction (proof in Appendix~\ref{app:atlas}), though tree consistency is an algebraic property of the graph rather than a guarantee that every local edge was estimated without error.

\paragraph{Gauge-instrumented trajectory benchmark and carry-forward baseline.}
We evaluate trajectory composition on a gauge-instrumented Qwen2.5-1.5B and Llama-3.2-1B fine-tuning benchmark spanning 9 runs per model (WikiText, code, math $\times$ seeds 42, 5042, 10042; 18 cross-run pairs) at 200 and 1,500 steps. At step 0 of each run, a sampled $\Bd$ relabeling is applied and recorded as ground truth, with checkpoints saved every 2 steps (100 local edges per chain at 200 steps; 750 at 1,500 steps). Because no further relabeling is injected during training, the step-0 map remains the true slot relation throughout training, and step-0 matching recovers it at 100.0\% on all 18 pairs (in-sample and held-out); retaining that initial map (the \emph{carry-forward baseline}) achieves 100.0\% without refitting. Moreover, a 1,500-step cross-run path uses 1,502 local fits (750 transitions plus one initial anchor per chain) and intermediate checkpoints, whereas direct endpoint matching uses 1 final-to-final fit. The comparison thus tests how step-wise local composition behaves relative to single-shot endpoint matching when intermediate checkpoints are available, rather than serving as an equal-compute replacement for retaining a known initial map.

\begin{table}[t]
\centering
\small
\setlength{\tabcolsep}{3.5pt}
\caption{Cross-run residual-stream coordinate recovery (\%, mean $\pm$ s.d.\ over 18 dependent pairs, higher is better; combined perm+sign differs by $\le 0.13\%$). Retained initial carries forward the step-0 map (100.0\%); Endpoint uses 1 final fit; Transport composes 2-step local $\Bd$ matches through the base (202 or 1,502 fits); Ep.\ wins counts pairs where Endpoint strictly exceeds Transport.}
\label{tab:cross-run}
\begin{tabular}{llrrrrr}
\toprule
& & \multicolumn{2}{c}{200 steps} & \multicolumn{2}{c}{1,500 steps} & \\
Model & Pair type / ablation & Endpoint & Transport & Endpoint & Transport & Ep.\ wins\\
\midrule
Qwen2.5-1.5B & Cross-seed (9 pairs) & $31.7\pm13.3$ & $96.2\pm6.6$ & $43.9\pm18.6$ & $91.0\pm8.4$ & 0/9, 0/9\\
& Cross-dataset (9 pairs) & $90.9\pm15.7$ & $96.2\pm6.4$ & $76.8\pm36.1$ & $91.2\pm9.6$ & 6/9, 6/9\\
& Overall (18 pairs) & $61.3\pm33.6$ & $96.2\pm6.3$ & $60.3\pm32.6$ & $91.1\pm8.7$ & 6/18, 6/18\\
\cmidrule{2-7}
& Endpoint-via-base & \multicolumn{2}{c}{$17.1\pm6.0$} & \multicolumn{2}{c}{$11.4\pm3.6$} & --\\
& Hybrid via base & \multicolumn{2}{c}{$37.6\pm11.1$} & \multicolumn{2}{c}{$26.7\pm6.0$} & --\\
& Retained initial & \multicolumn{2}{c}{$100.0\pm0.0$} & \multicolumn{2}{c}{$100.0\pm0.0$} & --\\
\midrule
Llama-3.2-1B & Overall (18 pairs) & $70.8\pm4.7$ & $88.6\pm6.0$ & $70.3\pm26.6$ & $66.9\pm37.1$ & 0/18, 9/18\\
& Retained initial & \multicolumn{2}{c}{$100.0\pm0.0$} & \multicolumn{2}{c}{$100.0\pm0.0$} & --\\
\bottomrule
\end{tabular}
\end{table}

\paragraph{Error compounding, horizon reversal, and retrospective localization.}
Table~\ref{tab:cross-run} shows both the strength of high-margin local steps and their vulnerability to long-horizon error compounding. On Qwen2.5-1.5B at 1,500 steps, 6,728/6,750 ($99.67\%$) of single-step transitions are recovered with 100\% exactness (22 local misses); composing 750 steps yields $95.3\%\pm 6.6\%$ within-run recovery ($98.1\%\pm 5.0\%$ at 200 steps; Table~\ref{tab:within-run}) and $91.1\%\pm 8.7\%$ cross-run recovery versus $60.3\%\pm 32.6\%$ for direct endpoint matching. Routing two endpoint maps through the base (\emph{Endpoint-via-base}) drops recovery to $11.4\%$, while direct endpoint matching wins on 6/9 cross-dataset Qwen pairs where shared-seed endpoint similarity is already near 100\% (Figure~\ref{fig:transport-advantage}). On Llama-3.2-1B, path composition beats direct endpoint matching by $+17.8$ points at 200 steps ($88.6\%$ vs.\ $70.8\%$, winning 9/9 chains and 18/18 pairs), but by 1,500 steps the ordering reverses to $66.9\%$ path versus $70.3\%$ endpoint ($-3.4$ points, with endpoint winning 9/18 pairs). Inspecting the local transition records localizes the cause: in the worst Llama chain (\texttt{wiki\_42}, which drops from $82.8\%$ at 200 steps to $0.29\%$ at 1,500 steps), two transitions six steps apart ($308\to 310$ and $314\to 316$) displace 1,901 and 2,017 of 2,048 coordinates with near-chance sign accuracy; although 728/750 total transitions (and 587/592 after step 316) are 100\% exact, group multiplication preserves the earlier corruption. In a row-budget-equalized Qwen2.5-0.5B 200-step panel and a smooth spacing sweep over strides $2$--$200$ (Tables~\ref{tab:budget-panel}--\ref{tab:spacing}, Appendix~\ref{app:trajectory-resources}), both endpoint and path stay at $96.2\%\text{--}100.0\%$ ($99.8\%\text{--}100.0\%$ at best spacing), with a 201-fit chain taking $5.6\text{--}7.8\,\mathrm{s}$ ($286.5\,\mathrm{MB}$ streaming) vs.\ $0.04\text{--}0.12\,\mathrm{s}$ for endpoint matching.

\section{Coordinate-preserving model merging}
\label{sec:merging}

When does linear weight interpolation require signed-permutation alignment rather than permutation-only or identity matching? In the coordinate-preserving regime, linear interpolation averages corresponding residual coordinates and RMSNorm signs are part of the correspondence. Following mode-connectivity evaluation, we measure the peak-above-chord loss barrier over $\alpha\in\mathcal A\subset[0,1]$, $\max_{\alpha\in\mathcal A}\{L(\alpha)-[(1-\alpha)L(0)+\alpha L(1)]\}$, isolating excess loss above the endpoint chord (lower is better).

\begin{table}[ht]
\centering
\small
\caption{Coordinate-preserving merge barriers ($\Delta\text{NLL}$ peak excess above the endpoint chord; lower is better). In the Llama-2-7B and TinyLlama rows, a known $\Bd$ relabeling is applied to one endpoint and $\Sd$ forces signs to $+1$; in the 10M independent-seed row, $\Sd$ and $\Bd$ are fitted from activations.}
\label{tab:merge}
\begin{tabular}{lrrr}
\toprule
Setting & Unaligned & $\Sd$ perm-only & $\Bd$ signed\\
\midrule
Llama-2-7B gauge-scramble + 50-step code FT & 6.14 & 7.69 & 0.00\\
10M RMSNorm, independent seeds, same data (WikiText) & 1.68 & 0.66 & 0.42\\
TinyLlama-1.1B, 0-step gauge-equivalent merge & 6.36 & 6.47 & 0.00\\
\bottomrule
\end{tabular}
\end{table}

Table~\ref{tab:merge} reports three controlled and small-scale barrier comparisons. On Llama-2-7B, we reparameterize the base model by a random $\Bd$ basis change, fine-tune the rebased copy for 50 steps on code, and interpolate weights between the original base and the aligned endpoint. Applying the true permutation without signs ($\Sd$ perm-only) causes destructive sign cancellation across opposite-sign coordinates (Proposition~\ref{prop:midpoint-cancellation}, Appendix~\ref{app:merge}), raising the barrier from 6.14 to 7.69, whereas signed $\Bd$ alignment eliminates it ($0.00$); the 0-step TinyLlama-1.1B merge shows the same separation ($6.36\to 6.47$ under $\Sd$ vs.\ $0.00$ under $\Bd$). In the 10M-parameter independent-seed experiment (two 4-layer, $d=256$ RMSNorm models trained from different seeds on WikiText for 1,000 steps without an imposed basis change), fitted $\Sd$ alignment reduces the barrier from 1.68 to 0.66 (61\% reduction), and fitted $\Bd$ reduces it to 0.42 (75\% reduction, 37\% relative improvement over $\Sd$; Appendix~\ref{app:merge}).

To test natural fine-tuned descendants at LLM scale, we fine-tune separate Qwen2.5-0.5B descendants on AG News and Yelp and compare their 50/50 weight average, task arithmetic ($\lambda\in\{0.25,0.5,0.75,1.0\}$), and TIES ($t\in\{0.2,0.5,0.8\}$; Table~\ref{tab:natural-merge-panel}, Appendix~\ref{app:natural-merge}). Because both descendants inherit the parent chart, their unmodified 50/50 average in inherited slots achieves 0.9119 AG News accuracy (0.5133 Yelp, 2.877 WikiText NLL), comparable to task arithmetic and TIES across the grid. When an exact $\Bd$ relabeling is planted on one descendant as a control, naive averaging collapses to 0.2700 AG News accuracy (13.025 NLL) and fitted $\Sd$ stays at 0.2494 (13.070 NLL), whereas fitted $\Bd$ restores 0.9125 accuracy (0.5233 Yelp, 2.880 NLL), matching gain-folded $O(d)$ (0.9056): inherited slots suffice for unmodified descendants, whereas $\Bd$ signs are essential under coordinate relabeling.

\section{Gauge audits expose coordinate non-reproducibility}
\label{sec:audits}

How can a pipeline that reports coordinate names be audited? Apply an exact symmetry of the relevant coordinate system, verify that model behavior is preserved, and compare raw versus mapped coordinate names (Table~\ref{tab:audits}). Residual dimensions, MLP hidden units, and SAE feature indices have distinct symmetries: a residual relabeling leaves MLP hidden activations and transported SAE feature activations unchanged, while permuting FFN hidden units or SAE feature indices is a separate symmetry. Across coordinate-indexed workflows, the represented functional object is invariant, but raw coordinate names drop to 0/2--0/10 overlap until the gauge map is applied.

\begin{table}[ht]
\centering
\small
\setlength{\tabcolsep}{4pt}
\caption{Coordinate-name audits under exact model reparameterizations (testing whether model function stays unchanged, raw coordinate names move, and mapping restores the named indices; lower error and higher overlap are better). BERT uses FFN permutations; others use residual $\Bd$.}
\label{tab:audits}
\begin{tabular}{llll}
\toprule
Pipeline & Invariance check & Raw overlap & After mapping\\
\midrule
BERT knowledge [Dai et al., 2022] & FFN error $3.24\times10^{-5}$ & 0/2 neurons & 2/2 neurons\\
Llama-2 refusal [Arditi et al., 2024] & logit error $3.36\times10^{-5}$ & median 0/10 & 10/10 coords\\
TinyLlama sycophancy [Rimsky et al., 2024] & effect error $2.85\times10^{-5}$ & mean 0/10 & effect preserved\\
Grad$\times$act attribution (4 decoders) & logit error $2.05\times10^{-5}$ & 0/10 indices & 10/10 indices\\
\bottomrule
\end{tabular}
\end{table}

\begin{proposition}[Sparse coordinate interventions force $\Bd$]
\label{prop:sparse}
For $Q\in O(d)$, the following are equivalent: (i) for every coordinate projector $P_S$, $QP_SQ^\top$ is a coordinate projector; (ii) for every $k=1,\ldots,d$, conjugation maps every $k$-coordinate projection edit to another; (iii) $Q\in\Bd$.
\end{proposition}

\paragraph{Proof sketch and scope.}
Signed permutations relabel coordinate projectors; conversely, applying (i) to singleton projectors $e_ie_i^\top$ forces each $Qe_i=\pm e_j$ with distinct targets by orthogonality, so $Q\in\Bd$. Thus dense rotations $Q\notin\Bd$ cannot substitute for $\Bd$ on sparse coordinate edits, as they turn a top-$k$ coordinate edit into a dense rank-$k$ projection (Appendices~\ref{app:related} and~\ref{app:limitations}).

\section{Conclusion}

In the native explicit-gain coordinate-preserving setting, RMSNorm transformers admit the signed-permutation gauge group $\Bd=\Sd\ltimes\{\pm1\}^d$ of order $2^d d!$, whereas LayerNorm transformers admit $\Sd\times C_2$ (Theorem~\ref{thm:native-gauge}). Omitting signs leaves signed-correlation matching with a $1/2+O(d2^{-d})$ ceiling under decorrelated coordinates (Corollary~\ref{thm:ceiling}) and introduces expected squared update error $2\norm{\Delta W}_F^2$ on weight deltas (Corollary~\ref{cor:sign-omission}), as verified by controlled reparameterizations across LoRA adapters, multi-behavior steering, merges, and interpretability audits. Across natural checkpoints, empirical validation defines the scope: inherited slots serve as the default for same-base descendants ($96.2\%\text{--}100\%$ retention, including $1.000$ self-match on \texttt{DeepSeek-R1-Distill-Qwen-1.5B}'s Math parent), multi-bank filters isolate sparse cross-generation subsets where negative signs causally govern held-out intervention effects, and trajectory composition recovers 91.1\% of cross-run Qwen coordinates at 1,500 steps while localizing step failures. Whenever a workflow names, edits, or transports RMSNorm coordinates, ``which neuron is this?'' is well-posed once $\Bd$ is specified.

\FloatBarrier
\clearpage
\section*{Acknowledgments and Disclosure of Funding}
Nothing to declare.

\section*{References}
\small
Samuel K. Ainsworth, Jonathan Hayase, and Siddhartha Srinivasa. Git Re-Basin: Merging models modulo permutation symmetries. In \emph{International Conference on Learning Representations}, 2023. URL \url{https://openreview.net/forum?id=CQsmMYmlP5T}.

Andy Arditi, Oscar Obeso, Aaquib Syed, Daniel Paleka, Nina Panickssery, Wes Gurnee, and Neel Nanda. Refusal in language models is mediated by a single direction. In \emph{Advances in Neural Information Processing Systems}, volume 37, 2024. URL \url{https://arxiv.org/abs/2406.11717}.

Saleh Ashkboos, Maximilian L. Croci, Marcelo Gennari do Nascimento, Torsten Hoefler, and James Hensman. SliceGPT: Compress large language models by deleting rows and columns. In \emph{International Conference on Learning Representations (ICLR)}, 2024a. URL \url{https://arxiv.org/abs/2401.15024}.

Saleh Ashkboos, Amirkeivan Mohtashami, Maximilian L. Croci, Bo Li, Pashmina Cameron, Martin Jaggi, Dan Alistarh, Torsten Hoefler, and James Hensman. QuaRot: Outlier-free 4-bit inference in rotated LLMs. In \emph{Advances in Neural Information Processing Systems}, volume 37, 2024b. doi: 10.52202/079017-3180. URL \url{https://proceedings.neurips.cc/paper_files/paper/2024/hash/b5b939436789f76f08b9d0da5e81af7c-Abstract-Conference.html}.

Johanni Brea, Berfin Simsek, Bernd Illing, and Wulfram Gerstner. Weight-space symmetry in deep networks gives rise to permutation saddles, connected by equal-loss valleys across the loss landscape. \emph{arXiv preprint arXiv:1907.02911}, 2019. URL \url{https://arxiv.org/abs/1907.02911}.

Trenton Bricken, Adly Templeton, Joshua Batson, Brian Chen, Adam Jermyn, Tom Conerly, Nick Turner, Cem Anil, Carson Denison, Amanda Askell, et al. Towards monosemanticity: Decomposing language models with dictionary learning. \emph{Transformer Circuits Thread}, 2023. URL \url{https://transformer-circuits.pub/2023/monosemantic-features/index.html}.

David F. Crouse. On implementing 2D rectangular assignment algorithms. \emph{IEEE Transactions on Aerospace and Electronic Systems}, 52(4):1679--1696, 2016. doi: 10.1109/TAES.2016.140952.

Damai Dai, Li Dong, Yaru Hao, Zhifang Sui, Baobao Chang, and Furu Wei. Knowledge neurons in pretrained transformers. In \emph{Proceedings of the 60th Annual Meeting of the Association for Computational Linguistics}, pages 8493--8502, 2022. doi: 10.18653/v1/2022.acl-long.581. URL \url{https://aclanthology.org/2022.acl-long.581/}.

Felix Draxler, Kambis Veschgini, Manfred Salmhofer, and Fred A. Hamprecht. Essentially no barriers in neural network energy landscape. In \emph{Proceedings of the 35th International Conference on Machine Learning}, volume 80 of \emph{Proceedings of Machine Learning Research}, pages 1309--1318, 2018. URL \url{https://proceedings.mlr.press/v80/draxler18a.html}.

Yanai Elazar, Nora Kassner, Shauli Ravfogel, Abhilasha Ravichander, Eduard Hovy, Hinrich Sch\"utze, and Yoav Goldberg. Measuring and improving consistency in pretrained language models. \emph{Transactions of the Association for Computational Linguistics}, 9:1012--1031, 2021. doi: 10.1162/tacl\_a\_00410. URL \url{https://aclanthology.org/2021.tacl-1.60/}.

Nelson Elhage, Robert Lasenby, and Christopher Olah. Privileged bases in the transformer residual stream. \emph{Transformer Circuits Thread}, 2023. URL \url{https://transformer-circuits.pub/2023/privileged-basis/index.html}.

Rahim Entezari, Hanie Sedghi, Olga Saukh, and Behnam Neyshabur. The role of permutation invariance in linear mode connectivity of neural networks. In \emph{International Conference on Learning Representations}, 2022. URL \url{https://openreview.net/forum?id=dNigytemkL}.

C. Daniel Freeman and Joan Bruna. Topology and geometry of half-rectified network optimization. In \emph{International Conference on Learning Representations}, 2017. doi: 10.48550/arXiv.1611.01540. URL \url{https://arxiv.org/abs/1611.01540}.

Leo Gao, Tom Dupr\'e la Tour, Henk Tillman, Gabriel Goh, Rajan Troll, Alec Radford, Ilya Sutskever, Jan Leike, and Jeffrey Wu. Scaling and evaluating sparse autoencoders. In \emph{International Conference on Learning Representations (ICLR)}, Oral, 2025. doi: 10.48550/arXiv.2406.04093. URL \url{https://openreview.net/forum?id=tcsZt9ZNKD}.

Timur Garipov, Pavel Izmailov, Dmitrii Podoprikhin, Dmitry Vetrov, and Andrew Gordon Wilson. Loss surfaces, mode connectivity, and fast ensembling of DNNs. In \emph{Advances in Neural Information Processing Systems}, 2018. doi: 10.48550/arXiv.1802.10026. URL \url{https://arxiv.org/abs/1802.10026}.

Robert Hecht-Nielsen. On the algebraic structure of feedforward network weight spaces. In \emph{Advanced Neural Computers}, pages 129--135. North-Holland, 1990. doi: 10.1016/B978-0-444-88400-8.50019-4. URL \url{https://doi.org/10.1016/B978-0-444-88400-8.50019-4}.

Gabriel Ilharco, Marco Tulio Ribeiro, Mitchell Wortsman, Suchin Gururangan, Ludwig Schmidt, Hannaneh Hajishirzi, and Ali Farhadi. Editing models with task arithmetic. In \emph{International Conference on Learning Representations}, 2023. URL \url{https://openreview.net/forum?id=6t0Kwf8-jrj}.

Moritz Imfeld, Jacopo Graldi, Marco Giordano, Thomas Hofmann, Sotiris Anagnostidis, and Sidak Pal Singh. Transformer fusion with optimal transport. In \emph{International Conference on Learning Representations}, 2024. doi: 10.48550/arXiv.2310.05719. URL \url{https://arxiv.org/abs/2310.05719}.

Roy Jonker and Anton Volgenant. A shortest augmenting path algorithm for dense and sparse linear assignment problems. \emph{Computing}, 38(4):325--340, 1987.

Simon Kornblith, Mohammad Norouzi, Honglak Lee, and Geoffrey Hinton. Similarity of neural network representations revisited. In \emph{Proceedings of the 36th International Conference on Machine Learning}, pages 3519--3529, 2019.

Harold W. Kuhn. The Hungarian method for the assignment problem. \emph{Naval Research Logistics Quarterly}, 2(1--2):83--97, 1955.

Yixuan Li, Jason Yosinski, Jeff Clune, Hod Lipson, and John Hopcroft. Convergent learning: Do different neural networks learn the same representations? In \emph{International Conference on Learning Representations}, 2016. URL \url{https://arxiv.org/abs/1511.07543}.

Stephen Merity, Caiming Xiong, James Bradbury, and Richard Socher. Pointer sentinel mixture models. In \emph{International Conference on Learning Representations}, 2017.

Ari Morcos, Maithra Raghu, and Samy Bengio. Insights on representational similarity in neural networks with canonical correlation. \emph{Advances in Neural Information Processing Systems}, 31, 2018.

Gon\c{c}alo Paulo and Nora Belrose. Sparse autoencoders trained on the same data learn different features. In \emph{International Conference on Learning Representations (ICLR)}, Poster, 2026. URL \url{https://openreview.net/forum?id=EjInprGpk9}.

Maithra Raghu, Justin Gilmer, Jason Yosinski, and Jascha Sohl-Dickstein. SVCCA: Singular vector canonical correlation analysis for deep learning dynamics and interpretability. \emph{Advances in Neural Information Processing Systems}, 30, 2017.

Nina Rimsky, Nick Gabrieli, Julian Schulz, Meg Tong, Evan Hubinger, and Alexander Turner. Steering Llama 2 via contrastive activation addition. In \emph{Proceedings of the 62nd Annual Meeting of the Association for Computational Linguistics (Volume 1: Long Papers)}, pages 15504--15522. Association for Computational Linguistics, 2024. doi: 10.18653/v1/2024.acl-long.828. URL \url{https://aclanthology.org/2024.acl-long.828/}.

Sidak Pal Singh and Martin Jaggi. Model fusion via optimal transport. In \emph{Advances in Neural Information Processing Systems}, 2020. URL \url{https://proceedings.neurips.cc/paper/2020/hash/fb2697869f56484404c8ceee2985b01d-Abstract.html}.

George Stoica, Daniel Bolya, Jakob Bjorner, Pratik Ramesh, Taylor Hearn, and Judy Hoffman. ZipIt! merging models from different tasks without training. In \emph{International Conference on Learning Representations}, 2024. URL \url{https://iclr.cc/virtual/2024/poster/18869}.

H\'ector J. Sussmann. Uniqueness of the weights for minimal feedforward nets with a given input-output map. \emph{Neural Networks}, 5(4):589--593, 1992. doi: 10.1016/S0893-6080(05)80037-1. URL \url{https://doi.org/10.1016/S0893-6080(05)80037-1}.

N. Joseph Tatro, Pin-Yu Chen, Payel Das, Igor Melnyk, Prasanna Sattigeri, and Rongjie Lai. Optimizing mode connectivity via neuron alignment. In \emph{Advances in Neural Information Processing Systems}, volume 33, pages 15300--15311, 2020.

Alexander Theus, Alessandro Cabodi, Sotiris Anagnostidis, Antonio Orvieto, Sidak Pal Singh, and Valentina Boeva. Generalized Linear Mode Connectivity for Transformers. In \emph{Advances in Neural Information Processing Systems}, 2025. URL \url{https://openreview.net/forum?id=KurYdcCbjv}.

Hugo Touvron, Louis Martin, Kevin Stone, Peter Albert, Amjad Almahairi, Yasmine Babaei, Nikolay Bashlykov, Soumya Batra, Prajjwal Bhargava, Shruti Bhosale, et al. Llama 2: Open foundation and fine-tuned chat models. \emph{arXiv preprint arXiv:2307.09288}, 2023. URL \url{https://arxiv.org/abs/2307.09288}.

Neha Verma and Maha Elbayad. Merging text transformer models from different initializations. \emph{Transactions on Machine Learning Research}, 2024. URL \url{https://openreview.net/forum?id=nWnYSLncXa}.

Pauli Virtanen, Ralf Gommers, Travis E. Oliphant, Matt Haberland, Tyler Reddy, David Cournapeau, Evgeni Burovski, Pearu Peterson, Warren Weckesser, Jonathan Bright, St\'efan J. van der Walt, Matthew Brett, Joshua Wilson, K. Jarrod Millman, Nikolay Mayorov, Andrew R. J. Nelson, Eric Jones, Robert Kern, Eric Larson, C. J. Carey, \.Ilhan Polat, Yu Feng, Eric W. Moore, Jake VanderPlas, Denis Laxalde, Josef Perktold, Robert Cimrman, Ian Henriksen, E. A. Quintero, Charles R. Harris, Anne M. Archibald, Ant\^onio H. Ribeiro, Fabian Pedregosa, Paul van Mulbregt, and SciPy 1.0 Contributors. SciPy 1.0: Fundamental algorithms for scientific computing in Python. \emph{Nature Methods}, 17:261--272, 2020. doi: 10.1038/s41592-019-0686-2.

Thomas Wolf, Lysandre Debut, Victor Sanh, Julien Chaumond, Clement Delangue, Anthony Moi, Pierric Cistac, Tim Rault, Remi Louf, Morgan Funtowicz, Joe Davison, Sam Shleifer, Patrick von Platen, Clara Ma, Yacine Jernite, Julien Plu, Canwen Xu, Teven Le Scao, Sylvain Gugger, Mariama Drame, Quentin Lhoest, and Alexander Rush. Transformers: State-of-the-art natural language processing. In \emph{Proceedings of the 2020 Conference on Empirical Methods in Natural Language Processing: System Demonstrations}, pages 38--45, Online, October 2020. Association for Computational Linguistics. doi: 10.18653/v1/2020.emnlp-demos.6. URL \url{https://aclanthology.org/2020.emnlp-demos.6/}.

Mitchell Wortsman, Gabriel Ilharco, Samir Y. Gadre, Rebecca Roelofs, Raphael Gontijo-Lopes, Ari S. Morcos, Hongseok Namkoong, Ali Farhadi, Yair Carmon, Simon Kornblith, and Ludwig Schmidt. Model soups: averaging weights of multiple fine-tuned models improves accuracy without increasing inference time. In \emph{Proceedings of the 39th International Conference on Machine Learning}, volume 162 of \emph{Proceedings of Machine Learning Research}, pages 23965--23998. PMLR, 2022.

Prateek Yadav, Derek Tam, Leshem Choshen, Colin Raffel, and Mohit Bansal. TIES-merging: Resolving interference when merging models. In \emph{Advances in Neural Information Processing Systems}, 2023. doi: 10.48550/arXiv.2306.01708. URL \url{https://arxiv.org/abs/2306.01708}.

An Yang, Baosong Yang, Beichen Zhang, Binyuan Hui, Bo Zheng, et al. Qwen2.5 technical report. \emph{arXiv preprint arXiv:2412.15115}, 2024a. URL \url{https://arxiv.org/abs/2412.15115}.

Enneng Yang, Li Shen, Guibing Guo, Xingwei Wang, Xiaochun Cao, Jie Zhang, and Dacheng Tao. Model merging in LLMs, MLLMs, and beyond: Methods, theories, applications and opportunities. \emph{arXiv preprint arXiv:2408.07666}, 2024b. URL \url{https://arxiv.org/abs/2408.07666}.

Biao Zhang and Rico Sennrich. Root mean square layer normalization. \emph{Advances in Neural Information Processing Systems}, 32, 2019.

Binchi Zhang, Zaiyi Zheng, Zhengzhang Chen, and Jundong Li. Beyond the permutation symmetry of transformers: The role of rotation for model fusion. In \emph{Proceedings of the 42nd International Conference on Machine Learning}, volume 267 of \emph{Proceedings of Machine Learning Research}, pages 77090--77106. PMLR, 2025. URL \url{https://proceedings.mlr.press/v267/zhang25dk.html}.

\normalsize

\clearpage
\appendix
\FloatBarrier
\section{Additional symmetry-boundary checks}
\label{app:figures}

\begin{figure}[ht]
\centering
\includegraphics[width=\linewidth]{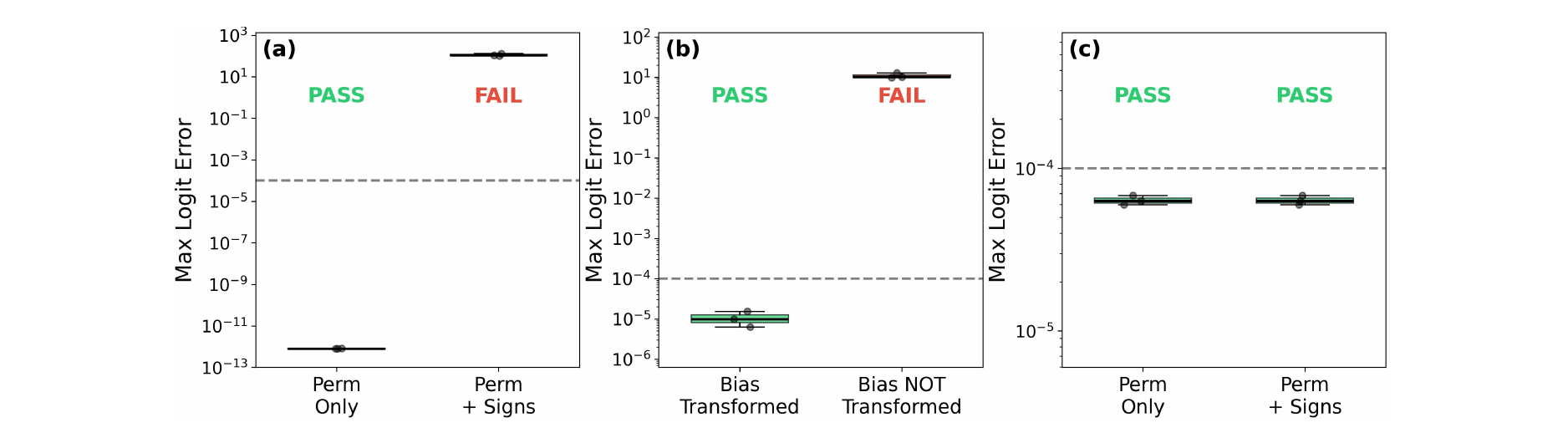}
\caption{Symmetry-boundary controls across three sampled gauges per arm (maximum absolute logit difference; lower is better). (a) GPT-2 with LayerNorm admits permutations but fails under independent signs. (b) TinyLlama with added layer-0 attention-output and MLP-output biases preserves logits only when those biases transform too. (c) Unmodified TinyLlama admits both permutations and signed permutations. Dots are individual gauges; boxes show interquartile ranges with median lines. Dashed lines mark $10^{-4}$.}
\label{fig:symmetry-boundary}
\end{figure}

\FloatBarrier
\section{Proof of the native-gauge theorem}
\label{app:proof-gauge}

We first establish module-level gauge transform rules (norms, linear layers, attention), and then compose them via residual addition and layerwise induction to prove Theorem~\ref{thm:native-gauge}. We present the decoder-only case for concreteness, using the row-vector convention: residual activations $H\in\R^{n\times d}$ are matrices with tokens as rows, and the gauge acts by right multiplication $H^G=HG$.

\begin{definition}[Gauge action]
For a permutation gauge $G=P\in\Sd$, $P$ is a permutation matrix with $P_{i,\pi(i)}=1$. For a signed-permutation gauge $G=PS\in\Bd$, we additionally have $S=\diag(s_1,\ldots,s_d)$ with $s_i\in\{-1,+1\}$. In both cases, the gauge action on residual activations $H\in\R^{n\times d}$ is $H^G:=HG$.
\end{definition}

\begin{lemma}[RMSNorm equivariance]
\label{lem:rmsnorm}
Define
\[
\RMSNorm(h;\gamma)=\frac{h}{\sqrt{\frac1d\sum_i h_i^2+\epsilon}}\odot\gamma
\]
for row vector $h$. With $\gamma^G_j:=\gamma_{\pi^{-1}(j)}$ (permute only, no sign), we have
\[
\RMSNorm(hG;\gamma^G)=\RMSNorm(h;\gamma)\cdot G.
\]
\end{lemma}

\paragraph{Proof.}
Mean square is invariant under signed permutations: $\frac1d\sum_i(hG)_i^2=\frac1d\sum_i h_i^2$. Thus $\RMS(hG)=\RMS(h)$, so $\RMSNorm(hG;\gamma^G)=hG/\RMS(h)\odot\gamma^G$. The permuted $\gamma^G$ aligns elementwise scaling with the permuted coordinates, giving $(h/\RMS(h)\odot\gamma)G$.

\paragraph{Why $\gamma$ permutes but does not flip signs.}
The sign flip is absorbed into the activation $h$, not $\gamma$. Specifically, $(hG)_j=s_jh_{\pi^{-1}(j)}$ carries the sign, while $\gamma^G_j=\gamma_{\pi^{-1}(j)}$ only permutes. The elementwise product therefore satisfies $(hG\odot\gamma^G)_j=s_jh_{\pi^{-1}(j)}\gamma_{\pi^{-1}(j)}=s_j(h\odot\gamma)_{\pi^{-1}(j)}=((h\odot\gamma)G)_j$.

\begin{lemma}[LayerNorm equivariance]
\label{lem:layernorm}
Define
\[
\LayerNorm(h;\gamma,\beta)=\gamma\odot\frac{h-\mu(h)}{\sigma(h)}+\beta,
\qquad
\mu(h)=\frac1d\sum_i h_i.
\]
For a permutation $P\in\Sd$ with $\gamma^P_j=\gamma_{\pi^{-1}(j)}$ and $\beta^P_j=\beta_{\pi^{-1}(j)}$,
\[
\LayerNorm(hP;\gamma^P,\beta^P)=\LayerNorm(h;\gamma,\beta)\cdot P.
\]
More generally, $\epsilon P$ with $\epsilon\in\{\pm1\}$ is equivariant if $\beta^P$ is replaced by $\epsilon\beta^P$. Signed permutations with a nonconstant sign pattern are not equivariant in general.
\end{lemma}

\paragraph{Proof.}
For a permutation $P$, $\mu(hP)=\frac1d\sum_j(hP)_j=\frac1d\sum_j h_{\pi^{-1}(j)}=\mu(h)$. Thus $(hP-\mu(hP))=(h-\mu(h))P$ and LayerNorm is $\Sd$-equivariant. For $\epsilon=-1$, $\mu(-hP)=-\mu(h)$ and the normalized centered vector changes by the same global sign, giving
\[
\LayerNorm(-hP;\gamma^P,-\beta^P)=\LayerNorm(h;\gamma,\beta)(-P).
\]
For a signed permutation with nonconstant sign flips, $\mu(hG)=\frac1d\sum_j s_jh_{\pi^{-1}(j)}\ne \pm\mu(h)$ in general. Thus $(hG-\mu(hG))\ne(h-\mu(h))G$, breaking equivariance.

\begin{lemma}[Post-norm wrapper]
\label{lem:postnorm}
Let $G$ be an admissible gauge. Suppose $\Delta$ is gauge-equivariant, $\Delta^G(HG)=\Delta(H)G$, and $N$ is gauge-equivariant, $N^G(HG)=N(H)G$. Then $\Phi(H)=N(H+\Delta(H))$ is gauge-equivariant: $\Phi^G(HG)=\Phi(H)G$.
\end{lemma}

\paragraph{Proof.}
$\Phi^G(HG)=N^G(HG+\Delta^G(HG))=N^G((H+\Delta(H))G)=N(H+\Delta(H))G$.

\begin{lemma}[Linear layer transforms under HuggingFace convention]
\label{lem:linear}
HuggingFace Linear computes $Y=XW^\top+b$ with $W\in\R^{d_{\rm out}\times d_{\rm in}}$. Under gauge $X^G=XG$:
\begin{itemize}
\item READ (input from residual, output unchanged): set $W^G=WG$, $b^G=b$. Then $Y^G=X^G(W^G)^\top+b^G=XGG^\top W^\top+b=XW^\top+b=Y$.
\item WRITE (output to residual): set $W^G=G^\top W$, $b^G=bG$. Then $Y^G=X(W^G)^\top+b^G=XW^\top G+bG=(XW^\top+b)G=YG$.
\end{itemize}
\end{lemma}

\begin{lemma}[Attention invariance]
\label{lem:attention}
With $Q,K,V$ projections as READ layers, the projected queries, keys, and values are unchanged under the gauge reparameterization. Therefore attention weights $\operatorname{softmax}(QK^\top/\sqrt{d_k})$ are identical, and only the output projection (WRITE) produces a gauged residual contribution.
\end{lemma}

\paragraph{Proof.}
Let $H^G=HG$ be the gauged input to attention. The Q-projection is a READ layer: $Q=HW_Q^\top+b_Q$ becomes
\[
Q^G=H^G(W_Q^G)^\top+b_Q=HGG^\top W_Q^\top+b_Q=HW_Q^\top+b_Q=Q.
\]
Similarly $K^G=K$ and $V^G=V$. The $Q/K/V$ biases live in head space, not residual space, so they remain unchanged under the gauge. Therefore attention weights are unchanged. The O-projection is WRITE, producing $\Delta^G=\Delta G$.

\paragraph{Proof of Theorem~\ref{thm:native-gauge}.}
We proceed by induction on layers, with $H_\ell^G=H_\ell G$ at layer $\ell$. In the base case ($\ell=0$), token embeddings $E=W_{\rm emb}[\mathrm{tokens}]$ are row vectors, so $W_{\rm emb}^G=W_{\rm emb}G$ gives $E^G=EG=H_0G$. For the inductive step, given $H_\ell^G=H_\ell G$, let $\Delta$ be the attention or MLP update. If the block is pre-norm, $H'=H_\ell+\Delta(N(H_\ell))$; Lemma~\ref{lem:rmsnorm} or~\ref{lem:layernorm} and Lemma~\ref{lem:attention} (or Lemma~\ref{lem:linear} for MLP) give $H'^G=H'G$. If post-norm, $H'=N(H_\ell+\Delta(H_\ell))$; $\Delta^G(H_\ell^G)=\Delta(H_\ell)G$ and Lemma~\ref{lem:postnorm} yields $H'^G=H'G$. Applying this to attention and MLP gives $H_{\ell+1}^G=H_{\ell+1}G$.

For the output, the LM head is READ from the final residual. With $W_{\rm head}^G=W_{\rm head}G$,
\[
\operatorname{logits}^G=H_L^G(W_{\rm head}^G)^\top
=H_LGG^\top W_{\rm head}^\top
=H_LW_{\rm head}^\top
=\operatorname{logits}.
\]
\emph{Tied embeddings:} When $W_{\rm head}=W_{\rm emb}$ (weight tying), both transform identically as $W^G=WG$, preserving the tie. Embedding rows and output-head rows both use right multiplication by the same $G$.

Two constraints determine the gauge group. For RMSNorm, the diagonal learned gain $\gamma$ must remain diagonal under the gauge. For a generic gain vector, or a jointly generic family of gain vectors sharing a residual chart, the simultaneous diagonalizers are exactly the signed permutation matrices $\Bd$. Exact tied coordinates across all relevant gains can admit block-orthogonal stabilizers, but these do not preserve individual coordinate axes; all results use the always-present $\Bd$ subgroup. For LayerNorm, mean-centering breaks independent sign symmetry. For a signed permutation $G$ with signs $s_i$, the mean $\mu(xG)=\frac1d\sum_j s_jx_{\pi^{-1}(j)}$ generally differs from $\pm\mu(x)$ unless all signs are equal. Thus generic affine LayerNorm admits $\{\pm P:P\in\Sd\}$, where the negative component requires $\beta\mapsto-\beta P$; its per-index relabeling content is $\Sd$. Both groups are large: $|\Sd|=d!$ and $|\Bd|=d!\cdot 2^d$. The factor of $2^d$ distinguishes the architectures, but both imply neuron indices are non-identifiable without fixing a gauge.

\begin{remark}[Joint gain degeneracies]
For RMSNorm with explicit diagonal gain, normalization covariance requires $G^\top\diag(\gamma^{(\ell)})G$ to remain diagonal for every normalization $\ell$ sharing the residual coordinates. A column of $G$ must therefore be a joint eigenvector of these diagonal matrices. Coordinates can mix continuously only when their tuples $(\gamma_i^{(\ell)})_\ell$ are identical. Ties in one normalization do not suffice; near ties do not give exact symmetry. Irrespective of these ties, $\Bd$ remains an admissible coordinate-preserving subgroup.

If a true correspondence were to contain a dense rotation within an exactly tied block, a signed-permutation fit would provide only a discrete approximation rather than recovering the continuous rotation, and gain degeneracy is separate from the finite-sample activation assignment margin of Lemma~\ref{lem:assignment-margin}. Appendix~\ref{app:gain-census} verifies on stored weights that all five inspected checkpoints have zero joint gain ties.
\end{remark}

\FloatBarrier
\section{Consistency of composed maps}
\label{app:atlas}

\paragraph{Proof of Proposition~\ref{prop:no-canonical}.}
Suppose $A:\Theta\times\Theta\to\G$ satisfies both function-level factoring, $A(\theta_1,\theta_2)=A(\theta_1,\theta_2H)$ for all $H\in\G$, and gauge-equivariance, $A(\theta_1,\theta_2H)=H^{-1}A(\theta_1,\theta_2)$ for all $H\in\G$. Combining the two identities gives $A(\theta_1,\theta_2)=H^{-1}A(\theta_1,\theta_2)$ for every $H\in\G$. Multiplying on the right by $A(\theta_1,\theta_2)^{-1}$ forces $H=I$ for every $H\in\G$, contradicting nontriviality of $\G$.

\paragraph{Proof of Proposition~\ref{prop:holonomy}.}
If $\hat G_{ij}=R_i^{-1}R_j$, then every cycle product telescopes:
\[
\hat G_{i_1i_2}\hat G_{i_2i_3}\cdots\hat G_{i_ki_1}
=R_{i_1}^{-1}R_{i_2}R_{i_2}^{-1}\cdots R_{i_k}^{-1}R_{i_1}
=I.
\]
Conversely, fix a root $r$. For any vertex $i$, choose a path from $r$ to $i$ and define $R_i$ as the ordered product of edge labels along that path. Zero holonomy makes $R_i$ path-independent: two paths from $r$ to $i$ form a cycle after reversing one path, so their products agree. For an edge $(i,j)$, the path to $i$ followed by that edge is a path to $j$; path-independence therefore gives $R_j=R_i\hat G_{ij}$, hence $\hat G_{ij}=R_i^{-1}R_j$.

A tree of fitted maps defines a consistent system by construction, because it has no independent cycles to test. This algebraic consistency does not establish accuracy against a recorded map or functional transfer between different models. Independent pairwise fits can be tested for cycle consistency; passing it is necessary for a common coordinate assignment, but is not sufficient evidence that the assignment is meaningful.

\FloatBarrier
\section{Proofs for artifact covariance}

\paragraph{Proof of Proposition~\ref{prop:update-covariance}.}
The parameter map $T_G$ is modulewise multiplication by fixed matrices $G$, $G^\top$, or by the corresponding permutation of norm gains. Hence it is linear in the base parameters and in their increments:
\[
T_G(\theta+\Delta\theta)-T_G(\theta)=T_G^{\rm lin}(\Delta\theta).
\]
For a LoRA update $\Delta W=BA$, the READ rule is $\Delta W^G=\Delta WG=B(AG)$, so $A^G=AG$, $B^G=B$. The WRITE rule is $\Delta W^G=G^\top\Delta W=(G^\top B)A$, so $A^G=A$, $B^G=G^\top B$.

\paragraph{Proof of Corollary~\ref{cor:sign-omission}.}
Write $G=PS$, with $S=\diag(s)$. Let $M=\Delta WP$. Then
\[
\norm{\Delta WPS-\Delta WP}_F^2
=\norm{MS-M}_F^2
=\sum_j(s_j-1)^2\norm{M_{:,j}}_2^2
=4\sum_{j:s_j=-1}\norm{(\Delta WP)_{:,j}}_2^2.
\]
Since $P$ is orthogonal, $\sum_j\norm{(\Delta WP)_{:,j}}_2^2=\norm{\Delta W}_F^2$. Under iid uniform signs, each column contributes with probability $1/2$, giving $2\norm{\Delta W}_F^2$ in expectation.

\paragraph{Proof of Lemma~\ref{lem:artifact-orientation}.}
The norm identity is the vector version of the preceding proof with $m=vP$:
\[
\norm{vG-vP}_2^2=\norm{mS-m}_2^2
=4\sum_{j:s_j=-1}m_j^2.
\]
For independent uniform signs, the weighted sign sum has mean zero and variance $\sum_j m_j^4/\|m\|_2^4$; Chebyshev's inequality gives concentration as this variance vanishes. The cosine follows from $\inner{vG}{vP}=\inner{mS}{m}=\sum_j s_jm_j^2$ and $\norm{vG}_2=\norm{vP}_2=\norm{m}_2$.

\FloatBarrier
\section{Decomposition and matching ceiling}
\label{app:decomp}

Theorem~\ref{thm:discrete-content} projects any known orthogonal matrix $Q\in O(d)$ onto its nearest signed permutation $P^\star\in\Bd$, with the residual rotation $R=Q(P^\star)^{-1}$ recording the continuous coordinate mixing left out by the discrete projection (whether $P^\star$ reflects a valid natural-checkpoint correspondence requires empirical validation, as in Section~\ref{sec:natural}).

\paragraph{Proof of Theorem~\ref{thm:discrete-content}.}
For any signed permutation $P\in\Bd$,
\[
\norm{Q-P}_F^2=\norm{Q}_F^2+\norm{P}_F^2-2\tr(Q^\top P)=2d-2\tr(Q^\top P),
\]
so minimizing Frobenius distance is equivalent to maximizing $\tr(Q^\top P)$. Write $P$ as a permutation $\sigma$ plus signs $\epsilon_j\in\{\pm1\}$, with column $j$ mapped to row $\sigma(j)$. Then
\[
\tr(Q^\top P)=\sum_j \epsilon_jQ_{\sigma(j),j}.
\]
For a fixed $\sigma$, the optimal sign is $\epsilon_j=\operatorname{sign}(Q_{\sigma(j),j})$, giving objective $\sum_j |Q_{\sigma(j),j}|$. Maximizing this over permutations is exactly a linear assignment problem on $|Q|$. Uniqueness holds if the maximizing assignment is unique and its matched entries are nonzero; it is not needed for the distance or factorization statements. Since both $Q$ and $P^\star$ are orthogonal, $R=Q(P^\star)^{-1}$ is orthogonal. Finally,
\[
\norm{R-I}_F=\norm{Q(P^\star)^{-1}-I}_F=\norm{Q-P^\star}_F
\]
by right-invariance of the Frobenius norm. The scalar $\rho(Q)=d^{-1}\tr(Q^\top P^\star)$ is one iff $Q=P^\star\in\Bd$ by the distance identity above. 

\paragraph{Proof of Corollary~\ref{thm:ceiling}.}
Let the population-centered coordinates satisfy $v_{\pi^\star(i)}=s_i^\star u_i$ up to sampling noise, and assume uncorrelated source coordinates $\mathbb{E}[u_iu_j]=0$ for $i\ne j$ with variance $\sigma_i^2>0$ on coordinate $i$. Assignment is unchanged by positive rescaling, so work with the empirical covariance $\widehat C=N^{-1}H_s^\top H_t$, which obeys
\[
\widehat C_{i,j}\xrightarrow[N\to\infty]{}s_i^\star\sigma_i^2\mathbf{1}\{j=\pi^\star(i)\}.
\]
All off-diagonal entries vanish in population. If $s_i^\star=+1$, the correct entry is the unique positive entry in row $i$, so signed-correlation maximization can select it. If $s_i^\star=-1$, the correct entry is the most negative population entry in its row, while all incorrect entries are zero; a maximizer of $\sum_i C_{i,\pi(i)}$ therefore avoids that entry whenever an off-diagonal assignment is available. Hungarian's global one-to-one constraint must displace some other row in exchange: when there are $k\ge2$ negative signs, deranging the $k$ negative-sign columns while fixing every positive-sign column attains the sum of all positive variances. This is an upper bound on every assignment objective, so every maximizer fixes all positive-sign rows and avoids all negative-sign entries. The event $k\le1$ has probability $O(d2^{-d})$ under iid uniform signs. Hence signed-correlation matching has limiting accuracy equal to the positive-sign fraction except on this event, giving expectation $1/2+O(d2^{-d})$ under independent uniform signs.

By contrast, $|\widehat C_{i,\pi^\star(i)}|\to\sigma_i^2>0$ and $|\widehat C_{i,j}|\to0$ for $j\ne\pi^\star(i)$. For fixed $d$, independent probe rows with finite second moments yield entrywise covariance convergence by the law of large numbers. Eventually the maximum entrywise perturbation is below $m/(2d)$, where $m>0$ is the population assignment gap; Lemma~\ref{lem:assignment-margin} then preserves the true assignment. Thus sign-marginalized matching converges to the true permutation.

\subsection{FFN hidden unit symmetry}

The residual gauge acts on the $d$-dimensional residual stream. MLP hidden units live in a separate $d_{\rm ffn}$-dimensional space. We first show that FFN hidden activations are invariant under the residual gauge, then prove they admit their own independent permutation symmetry.

\begin{lemma}[FFN hidden invariance under residual gauge]
\label{lem:ffn-hidden}
Under any admissible residual gauge $G$ (i.e., $G\in\{\pm P:P\in\Sd\}$ for LayerNorm models or $G\in\Bd$ for RMSNorm models), the FFN hidden-layer activations are invariant, not just equivariant. That is, if $H^G=HG$ and weights transform as READ/WRITE (Table~\ref{tab:module-rules}), then the $d_{\rm ffn}$-dimensional hidden activations are identical: $A^G=A$.
\end{lemma}

\paragraph{Proof.}
For a plain 2-layer FFN, let $\Delta_{\rm MLP}(H)=\sigma(HW_{\rm up}^\top+b_{\rm up})W_{\rm down}^\top+b_{\rm down}$ under row-vector convention. Under the residual gauge, $W_{\rm up}$ is READ, $W_{\rm up}^G=W_{\rm up}G$, $b_{\rm up}^G=b_{\rm up}$; $W_{\rm down}$ is WRITE, $W_{\rm down}^G=G^\top W_{\rm down}$, $b_{\rm down}^G=b_{\rm down}G$. The preactivation into the hidden units is
\[
Z^G=H^G(W_{\rm up}^G)^\top+b_{\rm up}^G
=(HG)(W_{\rm up}G)^\top+b_{\rm up}
=HG G^\top W_{\rm up}^\top+b_{\rm up}
=HW_{\rm up}^\top+b_{\rm up}=Z.
\]
Since $\sigma$ is elementwise, $A^G=\sigma(Z^G)=\sigma(Z)=A$. The hidden activations are \emph{literally identical}.

\emph{Gated MLP (Llama/SwiGLU).} For $\Delta_{\rm MLP}(H)=(\sigma(HW_{\rm gate}^\top+b_{\rm gate})\odot(HW_{\rm up}^\top+b_{\rm up}))W_{\rm down}^\top+b_{\rm down}$: both $W_{\rm gate}$ and $W_{\rm up}$ are READ layers, so the same $GG^\top=I$ cancellation applies to both branches. Thus both the gated and linear preactivations are unchanged, their Hadamard product is unchanged, and the hidden activations are invariant.

\emph{Output equivariance.} The down-projection is WRITE, so:
\[
\Delta_{\rm MLP}^G=A(W_{\rm down}^G)^\top+b_{\rm down}^G
=A(G^\top W_{\rm down})^\top+b_{\rm down}G
=AW_{\rm down}^\top G+b_{\rm down}G
=\Delta_{\rm MLP}\cdot G.
\]
The FFN contribution is equivariant (picks up $G$ on output), while internal hidden activations are invariant.

\begin{corollary}[FFN hidden indices require FFN gauge-fixing]
Methods that output top-$k$ FFN hidden unit indices (e.g., ``knowledge neurons'') are gauge-variant under the FFN symmetry ($S_{d_{\rm ffn}}$ for plain FFNs, $B_{d_{\rm ffn}}$ for gated FFNs), but are not affected by the residual gauge $G$. Thus such methods require gauge-fixing for the FFN symmetry, which is independent of the residual-stream symmetry analyzed in our main theorems.
\end{corollary}

We now formalize the internal hidden-unit symmetries of plain and gated FFN layers.

\begin{theorem}[FFN permutation symmetry]
For each FFN (MLP) layer with intermediate dimension $d_{\rm ffn}$ and elementwise activation $\sigma$, the permutation group $S_{d_{\rm ffn}}$ acts on the hidden unit indices while leaving the layer output invariant. Specifically, for $P\in S_{d_{\rm ffn}}$, define
\[
W'_{\rm up}=P^\top W_{\rm up},\quad b'_{\rm up}=b_{\rm up}P,\qquad
W'_{\rm down}=W_{\rm down}P,\quad b'_{\rm down}=b_{\rm down}.
\]
Then $\operatorname{FFN}'(x)=\operatorname{FFN}(x)$ for all inputs $x$. For gated FFNs (Llama/SwiGLU), the same permutation $P$ applies to both branches: $W'_{\rm gate}=P^\top W_{\rm gate}$ and $b'_{\rm gate}=b_{\rm gate}P$. For plain (non-gated) FFNs, sign flips are not a symmetry because common activations (GELU, SiLU, ReLU) are not odd functions. Gated FFNs admit additional sign symmetry; see Proposition~\ref{prop:gated-signed}.
\end{theorem}

\paragraph{Proof.}
Let $\operatorname{FFN}(x)=\sigma(xW_{\rm up}^\top+b_{\rm up})W_{\rm down}^\top+b_{\rm down}$ (row-vector convention), with $W_{\rm up}\in\R^{d_{\rm ffn}\times d_{\rm model}}$ and $W_{\rm down}\in\R^{d_{\rm model}\times d_{\rm ffn}}$.

Define the hidden activations $h:=\sigma(xW_{\rm up}^\top+b_{\rm up})\in\R^{1\times d_{\rm ffn}}$. With the transformed weights:
\[
(W'_{\rm up})^\top=(P^\top W_{\rm up})^\top=W_{\rm up}^\top P
\]
so the new preactivation is:
\[
x(W'_{\rm up})^\top+b'_{\rm up}=xW_{\rm up}^\top P+b_{\rm up}P=(xW_{\rm up}^\top+b_{\rm up})P.
\]
Since $\sigma$ is elementwise, it commutes with coordinate permutations:
\[
h'=\sigma((xW_{\rm up}^\top+b_{\rm up})P)=\sigma(xW_{\rm up}^\top+b_{\rm up})P=hP.
\]
For the output:
\[
(W'_{\rm down})^\top=(W_{\rm down}P)^\top=P^\top W_{\rm down}^\top
\]
so:
\[
y'=h'(W'_{\rm down})^\top+b_{\rm down}=(hP)(P^\top W_{\rm down}^\top)+b_{\rm down}=h(PP^\top)W_{\rm down}^\top+b_{\rm down}.
\]
Since $P$ is a permutation matrix, $PP^\top=I$, so $y'=hW_{\rm down}^\top+b_{\rm down}=y$.

\begin{proposition}[Signed-permutation subgroup of gated FFNs]
\label{prop:gated-signed}
For gated FFNs (SwiGLU/GLU as in Llama/Mistral/Qwen), $B_{d_{\rm ffn}}$ is an exact orthogonal hidden-coordinate symmetry subgroup, not merely $S_{d_{\rm ffn}}$. Specifically, for $P\in S_{d_{\rm ffn}}$ and $S=\diag(s_1,\ldots,s_{d_{\rm ffn}})$, define
\[
\begin{aligned}
&W'_{\rm gate}=P^\top W_{\rm gate},\quad b'_{\rm gate}=b_{\rm gate}P,\quad
W'_{\rm up}=SP^\top W_{\rm up},\quad b'_{\rm up}=b_{\rm up}PS,\\
&W'_{\rm down}=W_{\rm down}PS,\quad b'_{\rm down}=b_{\rm down}.
\end{aligned}
\]
Then $\operatorname{FFN}'(x)=\operatorname{FFN}(x)$ for all $x$, and the post-gating hidden vector transforms as $A'(x)=A(x)PS$.
\end{proposition}

\paragraph{Proof.}
Let $u(x)=xW_{\rm up}^\top+b_{\rm up}$ (linear branch) and $g(x)=xW_{\rm gate}^\top+b_{\rm gate}$ (gate branch). Under the transform:
\[
\begin{aligned}
u'(x)&=x(W'_{\rm up})^\top+b'_{\rm up}\\
&=x(SP^\top W_{\rm up})^\top+b_{\rm up}PS\\
&=(xW_{\rm up}^\top+b_{\rm up})PS=u(x)PS,
\end{aligned}
\]
and $g'(x)=g(x)P$, hence $\sigma(g'(x))=\sigma(g(x))P$ since $\sigma$ is elementwise. The post-gating hidden vector becomes:
\[
A'(x)=\sigma(g'(x))\odot u'(x)=(\sigma(g(x))P)\odot(u(x)PS)=(\sigma(g(x))\odot u(x))PS=A(x)PS.
\]
For the output:
\[
\begin{aligned}
\operatorname{FFN}'(x)&=A'(x)(W'_{\rm down})^\top+b_{\rm down}\\
&=A(x)PS(W_{\rm down}PS)^\top+b_{\rm down}\\
&=A(x)W_{\rm down}^\top+b_{\rm down}=\operatorname{FFN}(x),
\end{aligned}
\]
since $PS$ is orthogonal and $(PS)(PS)^\top=I$.

\begin{remark}[Additional diagonal scaling in gated FFNs]
The signed-permutation subgroup is not the full parameter symmetry of a gated FFN. For any invertible diagonal $D$ in hidden space, the transform $W'_{\rm up}=DW_{\rm up}$, $b'_{\rm up}=b_{\rm up}D$, $W'_{\rm down}=W_{\rm down}D^{-1}$, with the gate branch unchanged, also preserves the FFN function. This larger scaling symmetry is non-orthogonal unless $D$ has $\pm1$ diagonal entries, so it is separate from the residual-stream coordinate-preserving group studied in the main text.
\end{remark}

\begin{remark}[Why gated FFNs admit sign flips]
The key point is that the hidden-unit sign matrix acts on the \emph{linear} branch $u(x)$ (via $W_{\rm up}$), not inside the nonlinear gate $\sigma(g(x))$. Although $\sigma$ (SiLU/GELU) is not odd, the sign acts on $u(x)$ which is then multiplied by $\sigma(g(x))$. This is distinct from plain FFNs where the nonlinearity directly acts on the preactivation.
\end{remark}

\begin{remark}[Independence of symmetries]
The residual gauge $G\in\Bd$ and FFN hidden transform act on different index sets and can be applied simultaneously. The residual gauge acts on the $d_{\rm model}$ \emph{columns} of $W_{\rm up}$; the FFN hidden transform acts on the $d_{\rm ffn}$ \emph{rows}. For the signed hidden subgroup, write $K=PS$. The combined transform is:
\[
W_{\rm gate}^{G,P}=P^\top W_{\rm gate}G,\qquad
W_{\rm up}^{G,K}=K^\top W_{\rm up}G,\qquad
W_{\rm down}^{G,K}=G^\top W_{\rm down}K.
\]
Biases transform as $b_{\rm gate}\mapsto b_{\rm gate}P$, $b_{\rm up}\mapsto b_{\rm up}K$, and $b_{\rm down}\mapsto b_{\rm down}G$. These commute because $G$ and $K$ act on disjoint dimensions: $G$ permutes/signs the $d$-dimensional residual space while $K$ acts on the $d_{\rm ffn}$-dimensional hidden space.
\end{remark}

\begin{remark}[Why plain (non-gated) FFNs have no sign symmetry]
For plain 2-layer MLPs of the form $h=\sigma(xW_{\rm up}^\top+b)$, sign flips would require $\sigma(-z)=-\sigma(z)$ (oddness). $\operatorname{ReLU}(x)=\max(0,x)$ satisfies $\operatorname{ReLU}(-x)\ne-\operatorname{ReLU}(x)$ for $x>0$; similarly GELU and SiLU are not odd. Thus for \emph{plain} FFNs, sign flips do not preserve outputs, limiting these coordinate-preserving orthogonal hidden relabelings to $S_{d_{\rm ffn}}$. Plain ReLU FFNs also have the familiar positive diagonal scaling symmetry from ReLU homogeneity; GELU/SiLU plain FFNs generally do not. Gated FFNs (Proposition~\ref{prop:gated-signed}) admit this because the sign matrix acts on the linear branch while the nonlinear gate branch is only permuted.
\end{remark}

\subsection{SAE gauge invariance}
\label{app:sae}

Sparse autoencoders (SAEs) learn feature dictionaries from residual-stream activations. We show that the SAE training objective is invariant under residual gauge, while distinguishing a residual basis change from a permutation of learned feature indices.

\begin{proposition}[SAE objective is gauge-invariant]
\label{prop:sae-invariance}
Let $D=\{h_i\}_{i=1}^N\subset\R^d$ be residual activations (row-vector convention), and consider an SAE with encoder and decoder
\[
z(h)=\phi(hW_{\rm enc}^\top+b_{\rm enc}),\qquad
\hat h(h)=z(h)W_{\rm dec}^\top+b_{\rm dec},
\]
trained with objective
\[
L(\theta;D)=\sum_{h\in D}\norm{h-\hat h(h)}_2^2+\lambda\norm{z(h)}_1.
\]
Let $G$ be any admissible residual gauge (a permutation, possibly composed with the global sign flip, for LayerNorm models, or a signed permutation for RMSNorm models), and define $D^G=\{hG:h\in D\}$ and parameters
\[
W_{\rm enc}^G=W_{\rm enc}G,\quad b_{\rm enc}^G=b_{\rm enc},\quad
W_{\rm dec}^G=G^\top W_{\rm dec},\quad b_{\rm dec}^G=b_{\rm dec}G.
\]
Then for all $h\in D$,
\[
z^G(hG)=z(h),\qquad \hat h^G(hG)=\hat h(h)G,
\]
and consequently $L(\theta^G;D^G)=L(\theta;D)$.
\end{proposition}

\paragraph{Proof.}
Using $GG^\top=I$, $z^G(hG)=\phi(hGG^\top W_{\rm enc}^\top+b_{\rm enc})=z(h)$. Then $\hat h^G(hG)=z(h)W_{\rm dec}^\top G+b_{\rm dec}G=\hat h(h)G$. Thus $(hG-\hat h^G(hG))=(h-\hat h(h))G$, so the squared error is preserved because $G$ is orthogonal, and the $\ell_1$ term is unchanged because $z^G(hG)=z(h)$.

\begin{corollary}[SAE feature indices are non-identifiable]
Assume the feature activation $\phi$ commutes with feature permutations (as an elementwise nonlinearity does; TopK requires permutation-consistent tie handling). Independently of the residual gauge, for any feature permutation $P\in S_{d_{\rm feat}}$, the SAE objective is invariant under $W_{\rm enc}\mapsto P^\top W_{\rm enc}$, $b_{\rm enc}\mapsto b_{\rm enc}P$, and $W_{\rm dec}\mapsto W_{\rm dec}P$, with $b_{\rm dec}$ unchanged. Thus SAE feature indices are non-canonical without alignment.
\end{corollary}

\FloatBarrier
\section{Transform rules for all module types}
\label{app:rules}

All rules use the row-vector convention: $H^G=HG$ where $H$ has shape $[n,d]$. HuggingFace Linear computes $Y=XW^\top+b$ with $W\in\R^{d_{\rm out}\times d_{\rm in}}$.

\begin{table}[ht]
\centering
\small
\caption{Modulewise rules for the common residual gauge $H\mapsto HG$. READ outputs remain unchanged, WRITE outputs acquire $G$, and diagonal gains only permute.}
\label{tab:module-rules}
\begin{tabular}{llll}
\toprule
Module & Type & Weight & Bias\\
\midrule
Tok/Pos embed & Embed & $W\mapsto WG$ & --\\
Q, K, V proj & READ & $W\mapsto WG$ & unchanged\\
O projection & WRITE & $W\mapsto G^\top W$ & $b\mapsto bG$\\
MLP up/gate & READ & $W\mapsto WG$ & unchanged\\
MLP down & WRITE & $W\mapsto G^\top W$ & $b\mapsto bG$\\
RMSNorm & Norm & $\gamma\mapsto\gamma[\pi^{-1}]$ & --\\
LayerNorm & Norm & $\gamma\mapsto\gamma[\pi^{-1}]$ & $\beta\mapsto\epsilon\beta[\pi^{-1}]$ for $G=\epsilon P$; other signs break\\
LM head & READ & $W\mapsto WG$ & --\\
\bottomrule
\end{tabular}
\end{table}

In index form, let $\pi$ be the permutation with $P_{i,\pi(i)}=1$, and $s_j$ the signs. READ maps satisfy \mbox{$W^G[:,j]=s_jW[:,\pi^{-1}(j)]$} (permute/sign columns). WRITE maps satisfy \mbox{$W^G[i,:]=s_iW[\pi^{-1}(i),:]$} (permute/sign rows). WRITE bias transforms as $b_j^G=s_jb_{\pi^{-1}(j)}$. RMSNorm transforms as $\gamma_j^G=\gamma_{\pi^{-1}(j)}$ (permute only, no sign).

\paragraph{Why $Q,K,V$ are unchanged under reparameterization.}
This is the key insight that makes the proof work with multi-head attention. With $H^G=HG$ and $W_Q^G=W_QG$ (READ rule):
\[
Q^G=H^G(W_Q^G)^\top=HG\cdot G^\top W_Q^\top=HW_Q^\top=Q.
\]
The gauge cancels at the READ boundary, so attention patterns are identical. Table~\ref{tab:transport-rules} summarizes the induced $\Bd$ transport rules for downstream coordinate-indexed artifacts.

\begin{table}[ht]
\centering
\small
\caption{$\Bd$ transport rules for coordinate-indexed artifacts. $G\in\Bd$ acts on residual coordinates from the right ($H\mapsto HG$); READ matrices read from the residual, WRITE matrices write into it. Omitting signs loses orientation for vectors and first moments; second moments are sign-blind.}
\label{tab:transport-rules}
\begin{tabular}{lll}
\toprule
Artifact & Correct $\Bd$ transport & What $\Sd$ misses\\
\midrule
Steering / cached / probe vector $v$ & $v\mapsto vG$ & coordinate sign orientation\\
SAE dictionary $(W_{\rm enc},W_{\rm dec},b_{\rm dec})$ & $W_{\rm enc}G,\;G^\top W_{\rm dec},\;b_{\rm dec}G$ & read/reconstruction signs\\
LoRA READ / WRITE $\Delta W=BA$ & $(AG,B)$ / $(A,G^\top B)$ & input / output basis signs\\
First / second moment $(m_W,v_W)$ & $T_G(m_W),\;T_{|G|}(v_W)$ & $m_W$ signs ($v_W$ squares out)\\
\bottomrule
\end{tabular}
\end{table}

\FloatBarrier
\section{Hungarian optimality}

The sign-marginalized cost is optimal for recovering signed permutations. Let $u_i$ be column $i$ of $H$, $v_j$ column $j$ of $H^G$, and $C=H^\top H^G$. We seek a permutation $\pi$ and signs $s$ minimizing
\[
\min_{\pi\in\Sd,\;s\in\{\pm1\}^d}\sum_{i=1}^d\norm{u_i-s_iv_{\pi(i)}}_2^2.
\]
Expanding by columns,
\[
\sum_{i=1}^d\norm{u_i-s_iv_{\pi(i)}}_2^2
=\sum_i\left(\norm{u_i}_2^2+\norm{v_{\pi(i)}}_2^2-2s_i u_i^\top v_{\pi(i)}\right)
=\mathrm{const}-2\sum_i s_iC_{i,\pi(i)}.
\]
For any fixed permutation $\pi$, the optimal sign is $s_i=\operatorname{sign}(C_{i,\pi(i)})$, yielding
\[
\max_{\pi\in\Sd}\sum_i |C_{i,\pi(i)}|.
\]
This is exactly a linear assignment problem on cost $-|C_{i,j}|$, solvable by the Hungarian algorithm.

\subsection{Assignment stability under a margin}

\begin{lemma}[Assignment stability under a margin]
\label{lem:assignment-margin}
Let $C\in\R^{d\times d}$ and define $F_C(\pi)=\sum_{i=1}^d |C_{i,\pi(i)}|$ for $\pi\in\Sd$. Assume the maximizer $\pi^\star=\argmax_\pi F_C(\pi)$ is unique and define the margin
\[
m:=F_C(\pi^\star)-\max_{\pi\ne\pi^\star}F_C(\pi)>0.
\]
If $E\in\R^{d\times d}$ satisfies $\norm{E}_\infty=\max_{i,j}|E_{i,j}|\le\epsilon$ and $2d\epsilon<m$, then $\pi^\star$ remains the unique maximizer for $C+E$.
\end{lemma}

\paragraph{Proof.}
For any scalars $a,b$, $||a+b|-|a||\le |b|$. Thus for any $\pi$,
\[
|F_{C+E}(\pi)-F_C(\pi)|
\le\sum_{i=1}^d |E_{i,\pi(i)}|
\le d\epsilon.
\]
Therefore $F_{C+E}(\pi^\star)\ge F_C(\pi^\star)-d\epsilon$ and for any $\pi\ne\pi^\star$, $F_{C+E}(\pi)\le F_C(\pi)+d\epsilon\le F_C(\pi^\star)-m+d\epsilon$. Hence $F_{C+E}(\pi^\star)-F_{C+E}(\pi)\ge m-2d\epsilon>0$.

\FloatBarrier
\section{Experimental details}
\label{app:experiments}

\paragraph{Models.}
Our experiments span models from $\sim$10M to 8B parameters as summarized below, with the extended natural-checkpoint lineage and cross-generation panels detailed in Appendix~\ref{app:natural}. The symmetry-boundary checks of Table~\ref{tab:gauge-validation-full} cover seven models; the transport, LoRA, steering, and merging experiments use additional RMSNorm models.

\paragraph{Production-scale RMSNorm models.}
Qwen2.5-7B has 28 layers, $d=3584$, RMSNorm, and Q/K/V head-space biases only (boundary check). Llama-3.1-8B has 32 layers, $d=4096$, RMSNorm, and no biases (boundary check). Llama-2-7b-chat-hf has 32 layers, $d=4096$, RMSNorm, and no biases (boundary check, refusal reanalysis, gauge-scramble merge).

\paragraph{Mid-scale RMSNorm models.}
TinyLlama-1.1B has 22 layers, $d=2048$, RMSNorm, and no biases (boundary check, steering/LoRA transfer). Qwen2.5-1.5B has 28 layers, $d=1536$, RMSNorm, and Q/K/V head-space biases only (transport trajectory recovery, steering, LoRA, gauge-scramble). Qwen2.5-0.5B has 24 layers, $d=896$, RMSNorm, and Q/K/V head-space biases only (natural transfer and merging; Appendix~\ref{app:natural}). Llama-3.2-1B has 16 layers, $d=2048$, RMSNorm, and no biases (transport replication, steering replication; Appendix~\ref{app:transport-details}). Qwen Q/K/V biases live in attention head space and remain unchanged under the residual gauge, as in Lemma~\ref{lem:attention}; the reported transform rules do not rely on residual-space WRITE biases in these modules.

\paragraph{Small-scale and other architectures.}
We use a custom 10M Llama-style transformer with 4 layers, $d=256$, RMSNorm (independent-seed merging; trained from scratch on identical data, different seeds); BERT-base (110M, 12 layers, $d=768$, LayerNorm; $\Sd$ per-index symmetry; boundary check); T5-small (60M, 6+6 layers, $d=512$, T5LayerNorm; $\Bd$ symmetry; boundary check); and ViT-B/16 (86M, 12 layers, $d=768$, LayerNorm; $\Sd$ per-index symmetry; boundary check).

\paragraph{Existing assets and licenses.}
External checkpoints and datasets are obtained through their original distributions. Model weights are not redistributed; the refusal-replication code includes a small upstream prompt subset with its attribution and license notices. The Qwen2.5, TinyLlama, BERT-base, T5-small, and ViT checkpoints report Apache-2.0 license metadata on their Hugging Face model cards; Meta Llama checkpoints are gated and used under the corresponding Llama 2, Llama 3.1, or Llama 3.2 community license terms. WikiText is used under its CC-BY-SA-3.0/GFDL license metadata; SST-2/GLUE is used through the public GLUE distribution and original benchmark terms. ParaRel prompt templates used by the knowledge-neuron audit scripts are credited to Elazar et al. [2021]. Arditi et al.'s refusal-direction code and splits are credited in Appendix~\ref{app:arditi} and used only for the replication/audit reported there. Core software dependencies are PyTorch, Hugging Face Transformers/Datasets [Wolf et al., 2020], NumPy, and SciPy; their standard open-source licenses are BSD-style or Apache-2.0, and SciPy's assignment solver is cited where used.

\paragraph{Gauge transforms.}
We generate random gauges using NumPy with deterministic seeds: the permutation is a random permutation of $[0,d-1]$ with seed $s$; the signs are a random choice from $\{-1,+1\}^d$ with seed $s+1000$; multi-gauge sweeps use seeds 42--61 ($K=20$ gauges). All transforms are applied in-place to minimize memory overhead.

\paragraph{Steering vectors.}
We extract sycophancy steering vectors using Contrastive Activation Addition (CAA) by generating contrastive prompt pairs (sycophantic vs.\ honest responses), extracting residual-stream activations at layer 10, and computing the mean difference vector across prompt pairs. Steering effect is measured as the logit-difference change when adding the vector at inference.

\paragraph{Evaluation metrics.}
Logit invariance is evaluated on a fixed invariance batch (5 short prompts for language models; 5 random images for ViT), and we report the maximum absolute logit deviation across the batch. This batch is separate from the alignment probe set used for Hungarian gauge recovery. The logit error is the maximum absolute difference in output logits between original and gauge-transformed models, with threshold $\sim10^{-4}$ for gauge invariance. Top-$k$ overlap counts shared indices between top-$k$ coordinates by activation magnitude with $k=10$. Linear CKA uses standard centering across samples (HSIC with linear kernel), i.e.,
\[
\CKA(X,Y)=\frac{\norm{X_c^\top Y_c}_F^2}{\norm{X_c^\top X_c}_F\norm{Y_c^\top Y_c}_F},
\]
where $X_c,Y_c$ are centered across samples.

\paragraph{Refusal direction reanalysis (Arditi et al., 2024).}
\textbf{Setup.} We reproduce Arditi et al.'s released pipeline on \texttt{meta-llama/Llama-2-7b-chat-hf} (full protocol in Appendix~\ref{app:arditi}). \textbf{Candidates.} We form directions $r^{(\ell)}_i=\mathbb{E}[h^{(\ell)}_i\mid\mathrm{harmful}]-\mathbb{E}[h^{(\ell)}_i\mid\mathrm{harmless}]$ from block-input residual streams. \textbf{Selection.} We follow their bypass/induce criterion with KL filter 0.1 and layer cutoff $\ell<0.8L$; refusal score is last-token log-odds on $R=\{306\}$. \textbf{Checks.} On 20 harmful validation prompts (greedy decoding), ablating the selected direction reduces refusal from 20/20 to 3/20. We match their released direction artifact (layer 14, position $-1$) with cosine 0.984 in the same residual space ($d=4096$).

\paragraph{Hungarian alignment.}
The sign-marginalized recovery procedure uses SciPy's linear sum assignment solver [Virtanen et al., 2020]; SciPy documents this routine as a modified Jonker--Volgenant implementation described by Crouse [2016], building on shortest-augmenting-path assignment algorithms [Jonker and Volgenant, 1987]. Our gauge-derived cost matrices are near-permutation, and the dense assignment step completes in a few seconds for $d=4096$ in our implementation (worst-case $O(d^3)$). For $d\ge 8192$, we restrict each row to its top-$m$ matches in $|C|$ (we use $m=200$) and greedily match; this avoids cubic assignment but still forms the dense cross-correlation $C=H_{\mathrm{orig}}^{\top}H_{\mathrm{gauged}}$. In a $d=4096$ ablation, greedy top-200 matching recovers 94.7\% of the permutation and 97.5\% of the signs versus 100\% for exact assignment. Unless noted otherwise, the default alignment probe set consists of 30 diverse sentences totaling $\sim$500 tokens, and evaluation uses held-out test sentences disjoint from the probe set.

\paragraph{Controlled LoRA protocol.}
The controlled SST-2 experiments use rank 16, LoRA scale 32, dropout 0.05, learning rate $2\times10^{-4}$, batch size 4 with four-step gradient accumulation, maximum length 192, training seed 123, and gauge seed 456. Qwen2.5-1.5B uses 3,000 training examples and 800 steps; TinyLlama uses 1,000 examples and 250 steps. Their evaluation sets contain 872 and 200 examples, respectively. The Qwen configuration uses \texttt{bf16} weights and a 128-text alignment bank ($\le$20,000 tokens across four probe layers). Recovered and oracle $\Bd$ maps yield identical reported adapter metrics, and the remaining 0.945-versus-0.944 difference between the source and $\Bd$-transformed Qwen evaluations reflects a single validation example at finite precision.

\paragraph{Reproducibility.}
Unless otherwise stated, evaluation and gauge-recovery experiments load models in \texttt{float32} for numerical stability. The Qwen LoRA row uses a \texttt{bf16}-trained adapter; all raw, $\Sd$, $\Bd$, and reference evaluations for that row use the exact same trained adapter and checkpoint precision, isolating gauge transport rather than dtype. Transport scripts use fixed random seeds, request deterministic CUDA algorithms, and disable TF32.

\subsection{Compute resources}
\label{app:compute}

Experiments use single-worker PyTorch execution on CUDA GPUs (and Apple MPS with \texttt{float32} accumulators for small-model steering sweeps); no distributed training is required. Float32 7B/8B boundary and recovery checks use a single $\ge 48\,\mathrm{GB}$ GPU (an H200 worker was used for the 7B/8B gradient-times-activation attribution sweep), while all 1B--1.5B evaluation, LoRA, merging, and trajectory runs fit on a single 24\,GB-class GPU (100--300\,GB local disk recommended for Hugging Face caches and trajectory checkpoints). The four canonical 18-pair trajectory sweeps total 154.4 single-GPU hours (Table~\ref{tab:compute}).

\begin{table}[ht]
\centering
\small
\caption{Compute resources across experimental families, including measured wall-clock GPU times for the trajectory benchmarks and single-GPU memory footprints.}
\label{tab:compute}
\begin{tabular}{>{\raggedright\arraybackslash}p{0.24\linewidth}>{\raggedright\arraybackslash}p{0.38\linewidth}>{\raggedright\arraybackslash}p{0.26\linewidth}}
\toprule
Result family & Worker and memory & Run time / run size\\
\midrule
Symmetry-boundary, gauge-fixing, and attribution audits & Single CUDA worker. Float32 7B/8B rows use $\geq$48 GB VRAM (H200 for 7B/8B grad$\times$act attribution); smaller models fit on 24 GB GPUs. & 5--30 prompts per model; tens of minutes per 7B/8B checkpoint, dominated by model load and forward/backward passes.\\
\addlinespace
Qwen2.5-1.5B transport and steering transfer & Single CUDA worker, float32 ($\sim$6.2--6.3 GB VRAM allocated at model load; fits on one 24 GB GPU). & 22.8 h (200-step sweep) and 115.6 h (1,500-step sweep); each covers 9 trajectories and 18 cross-run pairs.\\
\addlinespace
Llama-3.2-1B transport replication & Single CUDA worker, float32 ($\sim$5.0--5.1 GB VRAM allocated at model load; fits on one 24 GB GPU). & 5.7 h (200-step sweep) and 10.3 h (1,500-step sweep); each covers 9 trajectories and 18 cross-run pairs.\\
\addlinespace
LoRA, SAE, natural-checkpoint, and steering transfers & Single CUDA worker (MPS/float32 for TinyLlama multi-behavior steering). Qwen LoRA uses rank 16, batch size 4, grad accum 4, 800 steps. & $\sim$1--3 GPU-hours per training/evaluation condition on a single 24 GB GPU.\\
\addlinespace
Merge barrier and natural merge panels & Single CUDA worker (24 GB GPU for 10M, Qwen0.5B, and TinyLlama; $\ge$48 GB for fp32 Llama-2-7B). & Small probe sets and 0--250 fine-tuning steps ($<1$ GPU-hour per merge panel).\\
\bottomrule
\end{tabular}
\end{table}

\subsection{Probe-budget sensitivity}
\label{app:probe}

This appendix collects probe-budget data referenced in Section~\ref{sec:recovery}. Exact reparameterization recovery, repeated path fitting, and natural-model stability are different evaluation regimes (Table~\ref{tab:probe-regimes}).

\begin{table}[ht]
\centering
\small
\setlength{\tabcolsep}{4pt}
\caption{Probe regimes across experimental families. Exact reparameterization recovery requires very few tokens; local trajectory composition reuses a fixed probe set across high-margin step transitions; natural cross-checkpoint comparisons require multi-bank prompt stability and held-out validation.}
\label{tab:probe-regimes}
\begin{tabular}{lll}
\toprule
Result family & Probe budget & Setting and validation criterion\\
\midrule
Recorded-gauge recovery & 22--500 tokens & exact $\Bd$ basis change (100\% recovery)\\
Controlled LoRA \& merge tests & $\sim$500--20K tokens & basis change + adapter or 0--50 step FT\\
Trajectory local composition & $\sim$500 tokens/edge & same-base fine-tuning chains (200--1500 steps)\\
Natural same-base descendants & 36K--99K tokens & Qwen2.5 Base/Math/Instruct, DeepSeek-R1-Distill\\
Natural cross-generation subsets & 3--4 prompt banks & WikiText/GSM8K/MBPP stability + held-out W2\\
\bottomrule
\end{tabular}
\end{table}

\textbf{Known-gauge exact-recovery regime: probe-robust.} On TinyLlama-1.1B with a sampled $\Bd$ gauge recorded as ground truth, sign-marginalized Hungarian recovery achieves 100\% permutation/sign/combined accuracy across all probe budgets we tested.

\begin{table}[ht]
\centering
\small
\caption{Probe-sensitivity on TinyLlama-1.1B with a sampled $\Bd$ gauge recorded as ground truth. Permutation, sign, and combined recovery are percentages (higher is better); all reach 100\% across the tested range.}
\label{tab:tiny-probe}
\begin{tabular}{rrrrr}
\toprule
Sentences & Tokens & Perm & Sign & Combined\\
\midrule
2 & 22 & 100\% & 100\% & 100\%\\
3 & 31 & 100\% & 100\% & 100\%\\
5 & 49 & 100\% & 100\% & 100\%\\
10 & 101 & 100\% & 100\% & 100\%\\
20 & 202 & 100\% & 100\% & 100\%\\
30 & 292 & 100\% & 100\% & 100\%\\
\bottomrule
\end{tabular}
\end{table}

\textbf{One same-base cross-dataset pair (Table~\ref{tab:qwen-probe}).} The Qwen2.5-1.5B probe sweep compares \texttt{wiki\_42} with \texttt{code\_5042} after 200 steps using a 10-step checkpoint stride against the recorded step-0 relabeling. Increasing the per-fit probe bank improves both single-shot endpoint and composed path recovery on this pair: at 7,873 tokens the endpoint reaches 99.35\% and the path reaches 100.00\%, and both reach 100.00\% at 33,678 tokens (note that this 200-step cross-dataset pair is easier than the 1,500-step cross-seed regime in Table~\ref{tab:cross-run}).

\begin{table}[ht]
\centering\small
\caption{Recorded permutation recovery (percent, higher is better) for one Qwen2.5-1.5B cross-dataset pair at 200 steps. Endpoint uses one fit; path composes ten-step-gap fits with the same bank per fit. Sign accuracy is 100\% in every arm, so combined accuracy equals permutation accuracy. The probe budgets are per fit, not matched totals.}
\label{tab:qwen-probe}
\begin{tabular}{lrrr}
\toprule
Probe set & Tokens & Endpoint & Path\\
\midrule
builtin\_5 & 39 & 28.13 & 47.46\\
builtin\_15 & 118 & 40.63 & 63.54\\
builtin\_35 & 270 & 41.08 & 63.61\\
wiki\_50 & 7,873 & 99.35 & 100.00\\
wiki\_200 & 33,678 & 100.00 & 100.00\\
\bottomrule
\end{tabular}
\end{table}

These curves are empirical finite-sample measurements: while Corollary~\ref{thm:ceiling} characterizes the $N\to\infty$ population limit rather than a finite-sample bound, Lemma~\ref{lem:assignment-margin} shows that unique recovery requires only a positive assignment margin (so low probe-matrix rank $n<d$ does not by itself preclude exact assignment).

\textbf{Natural cross-model regime: substantially probe-hungrier (Table~\ref{tab:natural-probe}).} For natural checkpoints without a known activation correspondence, one diagnostic is agreement with a large-probe reference fit. Here we compare Qwen2.5-1.5B Base against its same-base continued-pretraining descendant Qwen2.5-Math-1.5B (a probe-budget stability study, separate from the coordinate-preserving merge experiments in Section~\ref{sec:merging}); stability against the 36,010-token reference grows from 5.7\% at 39 tokens to 93.6\% at 27,044 tokens.

\begin{table}[ht]
\centering
\small
\caption{Natural-alignment stability vs.\ probe budget. Agreement is the fraction of coordinates assigned the same matched index as the wiki\_200 reference; sign-flip fraction is the share of matches with negative sign.}
\label{tab:natural-probe}
\begin{tabular}{lrrr}
\toprule
Probe set & Tokens & Agreement & Sign-flip frac.\\
\midrule
builtin\_5 & 39 & 5.7\% & 47.3\%\\
builtin\_15 & 118 & 11.5\% & 44.5\%\\
builtin\_35 & 270 & 15.7\% & 42.1\%\\
wiki\_10 & 1,830 & 73.2\% & 12.8\%\\
wiki\_50 & 6,619 & 82.9\% & 8.8\%\\
wiki\_100 & 18,032 & 89.4\% & 5.9\%\\
wiki\_150 & 27,044 & 93.6\% & 4.2\%\\
wiki\_200 (ref) & 36,010 & 100\% & 4.8\%\\
\bottomrule
\end{tabular}
\end{table}

\textbf{Practical scope.} Whereas the known-gauge cases tested here require only tens of tokens, natural-checkpoint fits vary with probe size and domain, and agreement with a larger reference fit measures estimation stability rather than ground-truth accuracy; natural pairs and long composed chains therefore require the empirical validation protocols of Sections~\ref{sec:natural} and~\ref{sec:trajectory}.

\subsection{Arditi et al. refusal direction replication}
\label{app:arditi}

\textbf{Code path.} We reproduce Arditi et al.'s released refusal-direction pipeline unmodified, adding a thin wrapper that contributes only the gauge analysis on top of their selection code.

\textbf{Prompts and splits.} We use Arditi's published harmful/harmless train and validation splits with \texttt{instructions\_only=true} and no shuffling. For Llama-2-7b-chat-hf, we use 128 harmful and 128 harmless train prompts to estimate mean activations, and 32 harmful plus 32 harmless validation prompts for direction selection. Behavioral refusal uses the first 20 harmful validation prompts.

\textbf{Prompt formatting.} Instructions are formatted with the Llama-2 chat template \texttt{[INST]\{instruction\}[/INST]} (no system prompt, trailing whitespace), matching Arditi's model class.

\textbf{Definition of $h$ and candidate directions.} We hook the transformer block input (residual stream before attention/MLP; pre-RMSNorm for Llama-2) using a forward pre-hook on each \texttt{model.model.layers[l]}. We record activations at the end-of-instruction token positions, i.e., the tokenized suffix \texttt{[/INST]} with indices $-|\mathrm{EOI}|,\dots,-1$. Candidate directions are $r_{\mathrm{pos},\ell}=\mathbb{E}[h_{\mathrm{pos},\ell}\mid\mathrm{harmful}]-\mathbb{E}[h_{\mathrm{pos},\ell}\mid\mathrm{harmless}]$, then L2-normalized.

\textbf{Direction selection (Arditi criteria).} For each $(\mathrm{pos},\ell)$ we compute (1) a refusal ablation score on harmful prompts using Arditi's global 3-site ablation (block input + attention output + MLP output at every layer), (2) a steering score on harmless prompts by adding the direction at the source layer only, and (3) a KL divergence score between baseline and ablated last-token logits on harmless prompts. We discard the last 20\% of layers, candidates with KL $>0.1$, and candidates with steering $<0$. We select the candidate with the lowest ablation refusal score, where refusal score is the log-odds of the model's refusal token(s) at the last prompt token (Llama-2 uses token id 306, ``I''). For this run the selected direction is layer 12, position $-2$.

\textbf{Cached direction comparison (cosine 0.984).} Arditi's released cached direction for Llama-2-7b-chat-hf is at layer 14, position $-1$. We compute cosine similarity between their cached vector and our candidate at the same layer/position (both L2-normalized in the same residual space, $d=4096$), obtaining 0.984. This compares candidate directions at the same layer and token position (layer 14, position $-1$), whereas automated selection on our validation split selects layer 12, position $-2$.

\textbf{Behavioral refusal rate ($100\%\to15\%$).} We generate completions for 20 harmful validation prompts with greedy decoding (\texttt{do\_sample=False}, \texttt{max\_new\_tokens=512}; no temperature/top-$p$). Refusal is the fraction of completions containing any of the following fixed substrings:
\begin{itemize}
\item \texttt{"i'm sorry"}, \texttt{"i am sorry"}, \texttt{"i apologize"}
\item \texttt{"as an ai"}, \texttt{"as a language model"}, \texttt{"as an assistant"}
\item \texttt{"i cannot"}, \texttt{"i can't"}, \texttt{"i am unable to"}, \texttt{"i'm unable to"}
\item \texttt{"i am not able to"}, \texttt{"i'm not able to"}
\end{itemize}
Baseline refusals are 20/20; global direction ablation reduces this to 3/20 (15\%).

\begin{table}[ht]
\centering
\small
\caption{Gauge invariance validation across architectures and scales. All models achieve max logit error $\lesssim 1.3\times10^{-4}$, consistent with the discrete gauge symmetry. LayerNorm models admit per-index $\Sd$ (plus the global sign flip); RMSNorm/T5LayerNorm models admit full $\Bd$ (signed permutations).}
\label{tab:gauge-validation-full}
\begin{tabular}{lllll}
\toprule
Model & Type & Norm & Gauge & Max error\\
\midrule
\multicolumn{5}{l}{\emph{Production-scale models (7--8B parameters)}}\\
Qwen2.5-7B & Decoder-only & RMSNorm & $\Bd$ & $1.01\times10^{-4}$\\
Llama-3.1-8B & Decoder-only & RMSNorm & $\Bd$ & $7.44\times10^{-5}$\\
Llama-2-7b-chat & Decoder-only & RMSNorm & $\Bd$ & $1.28\times10^{-4}$\\
\midrule
\multicolumn{5}{l}{\emph{Smaller models and other architectures}}\\
TinyLlama-1.1B & Decoder-only & RMSNorm & $\Bd$ & $8.58\times10^{-5}$\\
BERT-base & Encoder-only & LayerNorm & $\Sd$ & $2.29\times10^{-5}$\\
T5-small & Encoder-decoder & T5LayerNorm & $\Bd$ & $7.63\times10^{-5}$\\
ViT-B/16 & Vision & LayerNorm & $\Sd$ & $4.29\times10^{-6}$\\
\bottomrule
\end{tabular}
\end{table}

\begin{table}[ht]
\centering
\small
\caption{Signed-cost versus sign-marginalized Hungarian recovery under recorded $\Bd$ basis changes on 7B/8B RMSNorm models. All rows use layer 10, seed 42, and 30 probe texts. Signed-cost Hungarian maximizes raw correlations; sign-marginalized Hungarian matches by absolute correlation and then recovers the sign from the matched entry.}
\label{tab:signed-vs-abs}
\begin{tabular}{lrrrr}
\toprule
& \multicolumn{2}{c}{Signed-cost} & \multicolumn{2}{c}{Sign-marginalized}\\
Model & Perm. & Sign & Perm. & Sign\\
\midrule
Qwen2.5-7B & 49.4\% & 50.1\% & 100.0\% & 100.0\%\\
Llama-3.1-8B & 49.9\% & 50.1\% & 100.0\% & 100.0\%\\
Llama-2-7B-chat & 49.5\% & 50.1\% & 100.0\% & 100.0\%\\
\bottomrule
\end{tabular}
\end{table}

\FloatBarrier
\section{Multi-behavior steering results}
\label{app:steering}

\begin{table}[ht]
\centering
\small
\caption{Gauge equivariance across three steering behaviors (TinyLlama-1.1B-Chat; mean over all layers). Error (mean) is the mean absolute difference in baseline-subtracted steering effect (scale 2.0 minus 0.0), where steering effect is the mean per-token log-probability margin between positive and negative continuations. Top-10 overlap is the mean of $|\operatorname{top-10}(|v|)\cap\operatorname{top-10}(|v^G|)|$ across layers.}
\label{tab:steering-behaviors}
\begin{tabular}{llrr}
\toprule
Behavior & Extraction & Error (mean) & Top-10 overlap\\
\midrule
Sentiment & Raw text & $1.15\times10^{-5}$ & 0.05/10\\
Sycophancy & CAA & $2.85\times10^{-5}$ & 0.00/10\\
Refusal & Raw text & $1.98\times10^{-5}$ & 0.00/10\\
\bottomrule
\end{tabular}
\end{table}

We test gauge equivariance across three steering behaviors. For each, we extract steering vectors using either Contrastive Activation Addition (CAA) with chat-templated prompts, or raw contrastive text pairs. All three behaviors show the same pattern: steering effects are preserved (error $<10^{-4}$) while index-level summaries have near-zero overlap. Sentiment uses contrastive pairs of positive vs.\ negative affect statements. Sycophancy uses CAA prompts following Rimsky et al.\ [2024]: agreement-seeking vs.\ honest responses to opinion questions. Refusal uses contrastive pairs of compliant vs.\ refusing responses to benign requests.

\FloatBarrier
\section{Gauge tracking via parallel transport}
\label{app:transport-details}

We evaluate cross-run alignment by composing transports through a shared base checkpoint.

\paragraph{Setup.}
We fine-tune Qwen2.5-1.5B for 200 and 1,500 steps on three datasets (WikiText, code, math) with seeds 42, 5042, and 10042, yielding 9 runs and 18 cross-run pairs (unless noted otherwise, Appendix~\ref{app:transport-details} reports Qwen2.5-1.5B results, with Llama-3.2-1B summaries in Tables~\ref{tab:within-run} and~\ref{tab:steering-transfer}). At step 0 of each run, we apply a randomly sampled signed-permutation gauge and record the induced residual-stream correspondence as ground truth; carrying forward this initial map gives 100\% recovery on all 18 pairs at both horizons, including on held-out probes. To evaluate activation-based trajectory tracking, checkpoints are saved every two optimizer steps: 200-step runs use 101 checkpoints and 100 local edges per base-to-final chain, and 1,500-step runs use 751 checkpoints and 750 local edges, so a cross-run path composing two chains and two initial anchors uses 202 or 1,502 fits per pair. For each run we estimate a base-to-final map either by endpoint matching (a one-shot best-fit signed permutation at the endpoints) or by composing local gauges along the training trajectory (transport), and for each pair $(a,b)$ we compare direct endpoint matching against base-composition ablations (endpoint-via-base, hybrid) and transport-via-base. We report permutation recovery accuracy for compactness since permutation and combined (permutation+sign) accuracy differ by at most 0.13 percentage points across the Qwen pairs. The same fixed probe set is reused for each estimated edge; as a probe-overfitting check, direct endpoint accuracy on a held-out batch is lower by 2.8 percentage points on average at 1,500 steps.

\begin{figure}[ht]
\centering
\includegraphics[width=\linewidth]{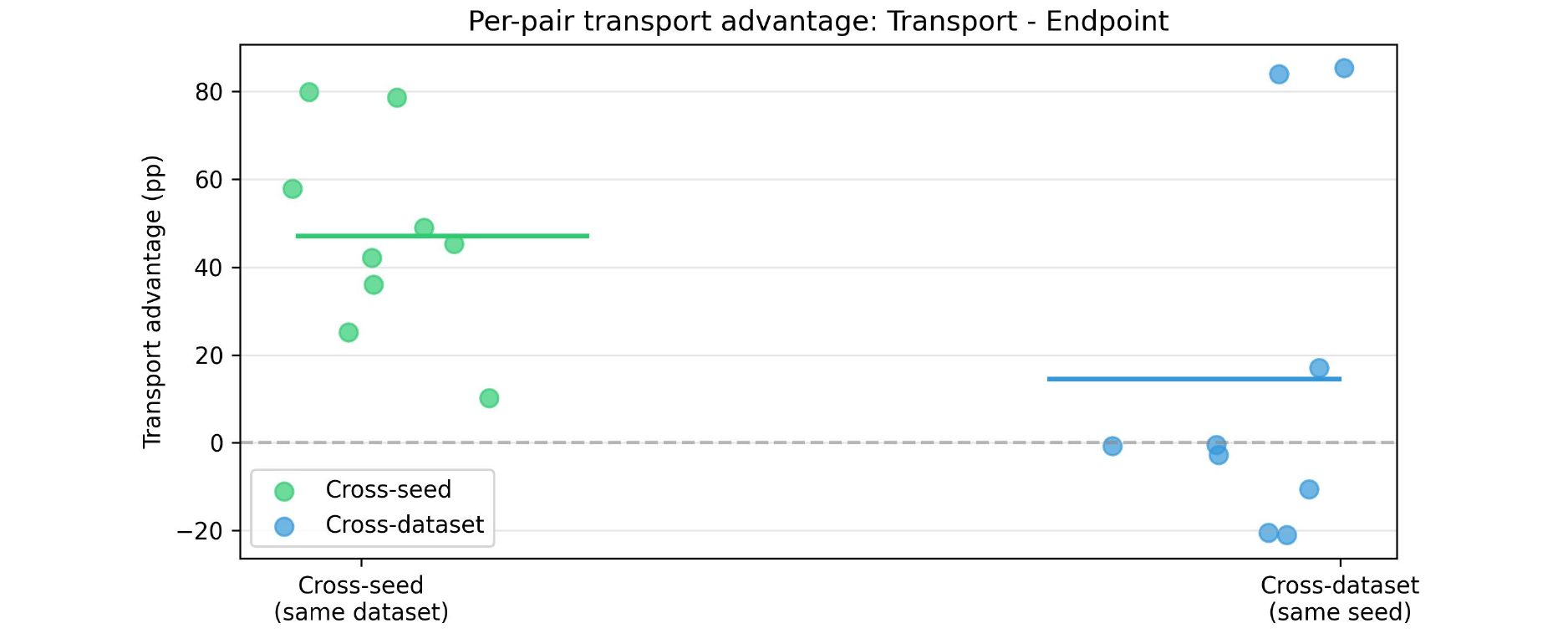}
\caption{Per-pair transport advantage $\Delta=\mathrm{Transport}-\mathrm{Endpoint}$ (percentage points) after 1,500 steps on Qwen2.5-1.5B. Each dot is one of nine cross-seed or nine cross-dataset pairs; horizontal bars denote group means. Transport consistently improves cross-seed recovery, while cross-dataset gains are heterogeneous and can be negative when endpoint matching is already near-ceiling.}
\label{fig:transport-advantage}
\end{figure}

\paragraph{Cross-run ablations and pair-type breakdown.}
Table~\ref{tab:cross-run} in Section~\ref{sec:trajectory} decomposes cross-run recovery into cross-seed and cross-dataset pairs for both Qwen2.5-1.5B and Llama-3.2-1B. Table~\ref{tab:cross-run-full} reports residual-stream cross-run permutation recovery on Qwen2.5-1.5B for five ways of constructing the cross-run map: direct final-to-final endpoint matching, composing endpoint gauges through the base (endpoint-via-base), two hybrid compositions (one side transport, one side endpoint), and composing transported gauges through the base (transport-via-base). Path fitting has the highest Qwen mean among these re-estimation arms at both horizons; the retained initial map is exact and is the stronger baseline in this benchmark. The largest transport gains over direct endpoint matching occur on cross-seed pairs, where direct endpoint matching is unreliable. Cross-dataset pairs are more heterogeneous: when endpoint matching is already near-ceiling, transport can underperform due to compounding local alignment errors (Figure~\ref{fig:transport-advantage}).

\begin{table}[ht]
\centering
\small
\caption{Residual-stream cross-run gauge recovery on Qwen2.5-1.5B. Endpoint-direct matches the two final checkpoints directly; the via-base and hybrid methods first align each run to the shared base (by endpoint matching or transport) and then compose through the base. Numbers report permutation recovery accuracy (\%, mean$\pm$sample s.d.) over 18 dependent run pairs from nine runs (3 datasets$\times$3 seeds).}
\label{tab:cross-run-full}
\begin{tabular}{lrr}
\toprule
Method & 200 steps & 1500 steps\\
\midrule
Endpoint-direct (final $\leftrightarrow$ final) & $61.3\pm33.6$ & $60.3\pm32.6$\\
Endpoint-via-base (endpoint$\cdot$endpoint) & $17.1\pm6.0$ & $11.4\pm3.6$\\
Hybrid (A transport, B endpoint) & $37.4\pm12.2$ & $26.9\pm6.0$\\
Hybrid (A endpoint, B transport) & $37.7\pm9.9$ & $26.5\pm6.1$\\
Transport-via-base (transport$\cdot$transport) & $96.2\pm6.3$ & $91.1\pm8.7$\\
\bottomrule
\end{tabular}
\end{table}

\textbf{Within-run recovery.} Table~\ref{tab:within-run} summarizes base$\to$final recovery within each run. Qwen residual transport has a 95.3\% mean at 1,500 steps, while FFN transport and individual residual chains can degrade; Table~\ref{tab:per-run} provides the per-run breakdown. We also include an additional within-run summary for Llama-3.2-1B.

\begin{table}[ht]
\centering
\small
\caption{Within-run gauge recovery (base $\to$ final) (percent, higher is better; mean $\pm$ sample s.d.) over 9 fine-tuning runs (3 datasets $\times$ 3 seeds). Transport composes local gauges along the training trajectory; endpoint matches activations only at the endpoints. FFN transport is evaluated only for Qwen2.5-1.5B.}
\label{tab:within-run}
\begin{tabular}{lrr}
\toprule
Metric & 200 steps & 1500 steps\\
\midrule
Qwen2.5-1.5B residual transport & $98.1\pm5.0$ & $95.3\pm6.6$\\
Qwen2.5-1.5B residual endpoint & $38.3\pm11.8$ & $27.9\pm6.4$\\
Qwen2.5-1.5B FFN transport & $96.3\pm4.0$ & $72.6\pm25.3$\\
Qwen2.5-1.5B FFN endpoint & $75.7\pm17.0$ & $30.7\pm6.7$\\
Llama-3.2-1B residual transport & $93.5\pm5.2$ & $81.6\pm31.2$\\
Llama-3.2-1B residual endpoint & $37.8\pm9.0$ & $20.9\pm8.0$\\
\bottomrule
\end{tabular}
\end{table}

\begin{table}[ht]
\centering
\small
\caption{Within-run permutation recovery (percent, higher is better) on Qwen2.5-1.5B after 1,500 fine-tuning steps. Residual and FFN maps act in different spaces; long-path FFN failures vary across runs.}
\label{tab:per-run}
\begin{tabular}{lrrrr}
\toprule
Dataset/seed & Res. endpoint & Res. transport & FFN endpoint & FFN transport\\
\midrule
code 10042 & 28.0 & 97.1 & 22.4 & 77.1\\
code 42 & 26.4 & 99.3 & 37.6 & 92.8\\
code 5042 & 25.7 & 99.2 & 29.7 & 97.5\\
math 10042 & 28.6 & 91.5 & 35.8 & 68.7\\
math 42 & 13.5 & 100.0 & 23.8 & 70.5\\
math 5042 & 32.1 & 97.5 & 42.5 & 13.3\\
wiki 10042 & 33.1 & 79.3 & 26.2 & 80.7\\
wiki 42 & 27.5 & 99.1 & 28.5 & 91.0\\
wiki 5042 & 35.9 & 95.1 & 30.3 & 61.9\\
\bottomrule
\end{tabular}
\end{table}

\subsection{Intervention transfer: cross-run steering vector transfer}

Gauge recovery accuracy measures coordinate-level alignment, but does this translate to downstream intervention transfer? We test whether sentiment steering vectors transfer between these separately fine-tuned same-base descendants under their recorded step-0 relabelings (rather than between independently pretrained models).

\textbf{Setup.} Using the same cross-run setup, we extract sentiment steering vectors and apply them across runs. We report results for Qwen2.5-1.5B (28 layers; residual layers $\{7,14,21\}$) and Llama-3.2-1B (16 layers; residual layers $\{4,8,12\}$). Let $T_r$ denote the transport base$\to$final map for run $r$ and $E_r$ the endpoint base$\to$final map (one-shot matching of $t_0$ and $t_T$). Then $G_{A\to B}=T_A^{-1}T_B$ and endpoint-via-base is $E_A^{-1}E_B$. For each pair of models $A\to B$, we compare:
\begin{itemize}
\item \textbf{Fresh:} Steering vector extracted directly from model $B$ (reference, not an upper bound)
\item \textbf{Transported:} Vector from $A$ transformed via transport gauge $G_{A\to B}$
\item \textbf{Endpoint-mapped:} Vector from $A$ transformed via endpoint-via-base mapping $E_A^{-1}E_B$
\end{itemize}
We measure steering effect preservation as $100\times\Delta_{\rm method}/\Delta_{\rm fresh}$. We treat a steering direction as detectable only if $\Delta_{\rm fresh}>0.1$; otherwise preservation is not reported and is excluded from averages.

\begin{table}[ht]
\centering
\small
\caption{Cross-run steering transfer after aligning source and target runs. We report Qwen2.5-1.5B and Llama-3.2-1B. Steering preservation is $100\times\Delta_{\rm method}/\Delta_{\rm fresh}$, where $\Delta$ is the baseline-subtracted steering effect (log-probability margin), and the win-rate is over detectable cases ($\Delta_{\rm fresh}>0.1$). For Qwen FFN we report only layer 14 (detectable in 14/18 pairs at 200 steps, 16/18 at 1500 steps). Values above 100\% indicate the transported vector exceeds the freshly extracted one; negative values indicate a reversed effect.}
\label{tab:steering-transfer}
\begin{tabular}{lrrr}
\toprule
Component & Transport & Endpoint & Win-rate\\
\midrule
Qwen residual, 200 steps & 103.6 & 33.2 & 54/54\\
Qwen residual, 1500 steps & 80.5 & 18.4 & 50/52\\
Qwen FFN layer 14, 200 steps & 84.8 & 25.2 & 13/14\\
Qwen FFN layer 14, 1500 steps & 38.3 & -6.2 & 12/16\\
Llama-3.2 residual, 200 steps & 99.1 & 53.5 & 54/54\\
Llama-3.2 residual, 1500 steps & 68.4 & 38.9 & 40/52\\
\bottomrule
\end{tabular}
\end{table}

\textbf{FFN steering detectability.} In the Qwen2.5-1.5B FFN steering-transfer evaluation we probe three \emph{pre-chosen} layers $\{7,14,21\}$ (for consistency with the residual-stream experiments). However, the fresh in-target FFN intervention is often negligible at layers 7 and 21, making preservation ratios ill-conditioned. We therefore treat a layer as \emph{not detectable} when $\Delta_{\rm fresh}\le 0.1$ and exclude it from preservation averages. Table~\ref{tab:ffn-detect} reports detectability rates for transparency: layer 14 is detectable in the majority of pairs, whereas layers 7 and 21 are rarely detectable.

\textbf{Reporting policy for FFN steering.} All FFN steering-transfer preservation numbers are reported for layer 14 only, where the fresh in-target intervention is most consistently detectable (Table~\ref{tab:ffn-detect}).

\textbf{Scope of steering transfer.} The cross-run steering-transfer experiment evaluates sentiment steering; broader behaviors and architectures are not evaluated in this experiment. FFN gauge tracking is less stable than residual-stream tracking over long trajectories in some runs (Table~\ref{tab:per-run}), suggesting more robust FFN matching (or periodic re-anchoring) is required for long-horizon FFN transport.

\begin{table}[ht]
\centering
\small
\caption{Detectability of FFN steering directions in the cross-run transfer evaluation (number of pairs with $|\Delta_{\rm fresh}|>0.1$; $0/18$ at layers 7 and 21 have signed $\Delta_{\rm fresh}>0.1$). Only layer 14 yields consistently detectable FFN steering.}
\label{tab:ffn-detect}
\begin{tabular}{lrr}
\toprule
FFN layer & 200 steps & 1500 steps\\
\midrule
7 & 3/18 & 0/18\\
14 & 14/18 & 16/18\\
21 & 3/18 & 3/18\\
\bottomrule
\end{tabular}
\end{table}

\FloatBarrier
\section{Additional merging experiments}
\label{app:merge}

\textbf{Question:} The merge experiments in Section~\ref{sec:merging} include gauge-scramble settings. Does sign-marginalized alignment also reduce interpolation barriers between independently initialized RMSNorm models?

\textbf{Observation:} In a small-scale independent-seed run, fitted signed alignment ($\Bd$) yields a lower interpolation barrier than fitted permutation-only alignment ($\Sd$), even though the raw negative-sign fraction alone does not certify coordinate correspondence.

\paragraph{Setup.}
We train two identical 4-layer Llama-style transformers ($d=256$, RMSNorm, $\sim$10M parameters) on WikiText [Merity et al., 2017] for 1,000 steps from different initialization seeds and fit $\Sd$ and $\Bd$ alignment maps via activation matching. Because this is a low-dimensional, same-data setting, its $\sim$500-token probe budget does not transfer to natural alignment of large LLMs, which remains probe-hungry as shown in Appendix~\ref{app:probe}.

\paragraph{Sign fractions vs.\ barrier reduction.}
Although roughly half of the pairs selected by absolute-correlation matching have negative correlation, the pairing-destroyed controls in Section~\ref{sec:natural} show that a $\sim 50\%$ negative-sign fraction can arise by chance without stable correspondence, so this fraction is a property of the fitted map rather than verified sign ground truth. We therefore evaluate downstream interpolation barriers in Table~\ref{tab:natural-merge}: permutation-only alignment ($\Sd$) reduces the peak barrier from 1.68 to 0.66 (61\% reduction), whereas signed alignment ($\Bd$) reduces it to 0.42 (75\% reduction, or a 37\% reduction in the remaining barrier), providing a small-scale merging comparison without claiming a full discrete correspondence between independently pretrained large models.

\begin{table}[ht]
\centering
\small
\caption{Merge barrier peaks for independently trained RMSNorm models (two independent seeds). The fitted signed map has a 37\% lower remaining barrier than the fitted permutation-only map. Barrier is peak excess NLL above the endpoint chord (lower is better).}
\label{tab:natural-merge}
\begin{tabular}{lrr}
\toprule
Alignment & Barrier peak ($\Delta$NLL) & Reduction\\
\midrule
Unaligned & 1.68 & --\\
Perm-only ($\Sd$) & 0.66 & 61\%\\
Signed ($\Bd$) & 0.42 & 75\%\\
\bottomrule
\end{tabular}
\end{table}

\subsection{Additional 0-step gauge-equivalent merge: TinyLlama-1.1B}

The same gauge-scramble merge setup also applies to a smaller RMSNorm decoder, \texttt{TinyLlama/TinyLlama-1.1B-Chat-v1.0}, with \textbf{0 fine-tuning steps}. Thus the two endpoints are \emph{exactly} gauge-equivalent, and a symmetry-respecting alignment should eliminate the peak barrier. Table~\ref{tab:tiny-merge} shows that perm-only alignment with the correct permutation but signs forced to $+1$ does not reduce the peak barrier relative to unaligned averaging, while signed alignment (ours; $\Bd$) eliminates it.

\begin{table}[ht]
\centering
\small
\caption{TinyLlama-1.1B gauge-scramble merge (0-step fine-tuning; endpoints are gauge-equivalent). Perm-only alignment with the correct permutation but signs forced to $+1$ leaves a large peak barrier, while signed alignment (ours; $\Bd$) eliminates it.}
\label{tab:tiny-merge}
\begin{tabular}{lr}
\toprule
Alignment & Barrier peak ($\Delta$NLL)\\
\midrule
Unaligned & 6.36\\
Perm-only (correct permutation) & 6.47\\
Signed (ours) & 0.00\\
\bottomrule
\end{tabular}
\end{table}

\subsection{\texorpdfstring{Minimal example: $\Sd$-only alignment can worsen averaging under $\Bd$}{Minimal example: permutation-only alignment can worsen averaging}}

We give a minimal example illustrating the cancellation mechanism behind ``perm-only worse than unaligned'' when the true symmetry group includes sign flips.

\begin{proposition}[Permutation-only alignment can increase midpoint loss under $\Bd$]
\label{prop:midpoint-cancellation}
There exist two $\Bd$-gauge-equivalent parameterizations of the same function such that averaging after a permutation-only ($\Sd$) alignment step yields strictly higher midpoint loss than unaligned averaging.
\end{proposition}

\paragraph{Proof.}
Consider the two-layer linear network $f_\theta(x)=xAB$ with $A=B=I_2$ and squared-error loss to targets $y=x$ under isotropic inputs with $0<\mathbb{E}\norm{x}_2^2<\infty$. Let
\[
P=\begin{bmatrix}0&1\\1&0\end{bmatrix},
\qquad
S=\diag(-1,1),
\qquad
G:=PS=\begin{bmatrix}0&1\\-1&0\end{bmatrix}\in\Bd,
\]
so $GG^\top=I_2$ and $G^\top=-G$. Define gauge-transformed parameters by $A^G=AG$ and $B^G=G^\top B$, so $f_{\theta^G}(x)=xA^GB^G=x$ and $\theta$ and $\theta^G$ are gauge-equivalent.

For unaligned averaging, midpoint parameters are $A_{1/2}=\frac12(I_2+G)$ and $B_{1/2}=\frac12(I_2+G^\top)$, giving $A_{1/2}B_{1/2}=\frac14(I_2+G+G^\top+GG^\top)=\frac12I_2$. Thus $f_{1/2}(x)=\frac12x$ and expected midpoint loss is $\frac14\mathbb{E}\norm{x}_2^2$.

For $\Sd$-only alignment, a permutation-only alignment can remove the permutation component $P$ but cannot apply the required sign flip. Composing with $P$ yields sign-only mismatch $GP=\diag(1,-1)$. Averaging $\theta$ with $\theta^{GP}$ gives $A_{1/2}=\frac12(I_2+GP)=\diag(1,0)$ and $B_{1/2}=\diag(1,0)$, so $f_{1/2}(x)=(x_1,0)$. The expected midpoint loss is $\mathbb{E}[x_2^2]=\frac12\mathbb{E}\norm{x}_2^2$, strictly larger than $\frac14\mathbb{E}\norm{x}_2^2$.

\FloatBarrier
\section{Extended related work}
\label{app:related}

\paragraph{Permutation symmetries and mode connectivity.}
Hidden-unit permutations have long been known to create equivalent parameterizations and connected loss landscapes [Brea et al., 2019, Freeman and Bruna, 2017, Draxler et al., 2018, Garipov et al., 2018, Tatro et al., 2020, Entezari et al., 2022]. Git Re-Basin turns this into an alignment procedure over $\Sd$ [Ainsworth et al., 2023]. Our contribution is the architecture-dependent residual-stream group for transformers: per-index $\Sd$ for LayerNorm and $\Bd$ for RMSNorm, together with the decorrelated-coordinate corollary illustrating the failure of a sign-blind assignment objective.

\paragraph{Continuous and rotation-based methods.}
SliceGPT, QuaRot, generalized linear mode connectivity, and rotation-aware fusion exploit broader $O(d)$ freedom after reparameterization [Ashkboos et al., 2024a,b, Theus et al., 2025, Zhang et al., 2025]. These methods address a different alignment regime: they can be stronger for coordinate-indifferent compression or fusion, but they do not provide coordinate-preserving transport of sparse objects in the native $\gamma$-explicit parameterization. Theorem~\ref{thm:discrete-content} gives the interface: a nearest $\Bd$ approximation preserves coordinate axes, but its empirical quality still requires validation.

\paragraph{Model merging and interpretability.}
Merging methods include weight matching, model soups, task arithmetic, TIES, ZipIt!, and optimal-transport fusion [Ainsworth et al., 2023, Wortsman et al., 2022, Ilharco et al., 2023, Yadav et al., 2023, Stoica et al., 2024, Singh and Jaggi, 2020, Imfeld et al., 2024, Verma and Elbayad, 2024, Yang et al., 2024b]. Our merge results are not a universal merging recipe; they identify a necessary discrete correction for RMSNorm coordinate-preserving alignment.

In interpretability, invariant representation metrics remain valid without gauge fixing [Raghu et al., 2017, Morcos et al., 2018, Kornblith et al., 2019]; direction and subspace analyses can be compared after consistent coordinate transport [Elhage et al., 2023]. Our audit targets the complementary class of analyses that output indices. SAE dictionary studies identify feature indices with semantic labels [Bricken et al., 2023, Gao et al., 2025]: Bricken et al. [2023] report ``feature A/1/2357'' as a base64-input detector and characterize hundreds of similarly indexed features with concept labels. Knowledge-neuron pipelines locate factual associations at specific MLP hidden units: Dai et al. [2022] identify $w^{(9)}_{2141}$ and $w^{(10)}_{1122}$ as carrying the Ireland--Dublin association, and report similar coordinate triples for other facts. Direction-level steering work [Rimsky et al., 2024, Arditi et al., 2024] makes single-model claims that are gauge-invariant in isolation but require a coordinate map to be compared or transferred across runs or related models, whose exact transport follows the vector rule in Lemma~\ref{lem:artifact-orientation}.

These workflows also involve distinct native symmetries: BERT-style FFN neuron labels change under hidden-unit permutations, whereas an SAE's feature-index permutation is independent of the residual basis because transporting the encoder and decoder together leaves feature activations unchanged (Proposition~\ref{prop:sae-invariance}). Moreover, because cross-seed SAE disagreements can arise from optimization, data order, and feature splitting [Paulo and Belrose, 2026], a residual-coordinate transport rule isolates basis changes but cannot by itself identify which of these mechanisms caused a disagreement.

\FloatBarrier
\section{Natural-checkpoint protocols and additional results}
\label{app:natural}

The experiments in this appendix evaluate unmodified public checkpoints---Qwen2 and Qwen2.5 at 0.5B and 1.5B (including Qwen2.5-Math-1.5B, Instruct, and Coder), SmolLM2-360M Base/Instruct, Llama-3.2-1B Base/Instruct, and the TinyLlama variants specified below---except where an explicit relabeling is introduced as a control. Because naturally trained checkpoints compute different functions, a fitted map represents an empirical coordinate correspondence rather than proof of functional equivalence, and measurements across multiple descendants of a single parent provide dependent evidence about that lineage.

\subsection{Inherited slots, phylogeny, and natural artifact transfer}

The Qwen2.5-1.5B ancestry analysis fits maps from paired residual activations pooled across layers 7, 14, and 21. In its pooled Base/Math fit, 96.224\% of indices retain their original partner while the median absolute matched correlation is 0.300, showing that coordinate-slot retention is distinct from raw activation-value similarity. Similarly, Base/Instruct fits retain every index across Qwen2.5-0.5B, SmolLM2-360M, and Llama-3.2-1B, and comparing TinyLlama v1.1 against its Math/Code and Chinese continued-pretraining variants likewise yields identity assignments with $+1$ signs (with mean absolute diagonal correlations of 0.874 and 0.823). Extending the 1.5B lineage comparison to \texttt{DeepSeek-R1-Distill-Qwen-1.5B} across all five 1.5B checkpoints (\texttt{Qwen2.5-1.5B} Base, Instruct, Coder, Math, and \texttt{Qwen2-1.5B}) recovers its exact parentage by self-matched coordinate rate and cross-bank agreement: R1 matches \texttt{Qwen2.5-Math-1.5B} at $1.000$ self-match ($1.000$ cross-bank agreement, $0.694$ same-index correlation, $\rho(Q)=0.966$), versus $0.915$ against Coder ($0.871$ cross-bank), $0.903$ against Instruct ($0.846$ cross-bank), $0.896$ against Base ($0.840$ cross-bank), and $0.002$ against cross-generation \texttt{Qwen2-1.5B} ($0.035$ cross-bank, $\rho(Q)=0.110$ vs.\ null $0.088$). These empirical findings characterize how standard fine-tuning preserves inherited slots in practice rather than a theorem that an optimizer cannot exchange coordinate roles.

\paragraph{LoRA and steering.}
The natural Base/Math experiment trains one rank-16 SST-2 LoRA adapter on each model for 800 steps (seed 4242, batch size 16, learning rate $10^{-4}$, maximum length 128, \texttt{float32}), targeting attention Q/K/V/O projections. The same trained adapter is evaluated under every transfer map on the 872-example SST-2 validation set, with Signed ($\Bd$), permutation-only ($\Sd$), and Dense $O(d)$ maps fitted on disjoint WikiText-103 activations under their respective objectives. The unadapted Base and Math target accuracies are 0.549 and 0.630 in this setup (distinct from the controlled \texttt{bf16} LoRA setup in Table~\ref{tab:artifact-covariance}).

Sentiment steering uses a contrastive direction at layer 14, with extraction prompts disjoint from evaluation prompts. The steering effect is the shift in the positive-versus-negative continuation log-probability margin (fresh target effects are 1.716 on Math and 1.433 on Base; Table~\ref{tab:natural-transfer} reports the transferred effects).

\paragraph{SAE transfer.}
The natural Base$\to$Math SAE transfer experiment trains a layer-14 TopK SAE (6,144 features, $k=32$, 3,000 optimization steps, batch size 4,096, learning rate $10^{-3}$) on 99,328 tokens across W1, GSM8K, and MBPP, and evaluates reconstruction on 35,840 held-out W2 tokens. Alignment maps are fitted on W1 only. Activations receive a scalar normalization from their training split, and NMSE is total squared reconstruction error divided by total squared deviation from the target mean (excluding the first two token positions per 1,024-token sequence and the top 1\% of activation norms). The source SAE achieves 0.301 NMSE on its source model and a freshly trained target SAE achieves 0.329 on the target model. Because each map and its exact inverse wrap the encoder and decoder, an identity dictionary reconstructs identically across arms, isolating dictionary compatibility from the linear wrapper. Standard dense orthogonal Procrustes alignment yields 1.115 NMSE (a whiten/rotate/unwhiten variant yields 5.210 due to ill-conditioned whitening scales).

\subsection{Prompt selection and intervention evaluation}

For Qwen2-0.5B versus Qwen2.5-0.5B, W1 and W2 are disjoint WikiText-103 splits, M is GSM8K, and C is MBPP. Matching uses zero-based block-output layers 6, 12, and 18. Each layer's correlation matrix is computed after dropping the first two token positions per sequence, regressing each coordinate on an intercept and the activation-vector $\ell_2$ norm, and standardizing its residual; correlation matrices are then averaged across the three depths. Selection requires exact partner and sign agreement across W1, M, and C, holding W2 out strictly for causal intervention evaluation. This three-bank filter retains 28 coordinate pairs (10 with negative signs).

Interventions are injected at every token at block output 12: a source coordinate receives $+a\sigma_s$ and its matched target coordinate receives $a s\sigma_t$, using the fitted sign $s\in\{\pm1\}$ and coordinate-specific standard deviations measured on W1 after initial-position removal. The sign-removal control applies $+a\sigma_t$ at the same target partner, while the separately fitted $\Sd$ arm maximizes unsigned correlation and may select a different partner coordinate. Evaluation measures the Pearson correlation between source and target output-logit shifts across evaluation tokens and 4,096 sampled vocabulary entries on 24 held-out W2 sequences, averaged across selected coordinate pairs.

The dense $O(d)$ comparator fits an orthogonal Procrustes map on standardized W1 activations, maps a unit source coordinate to the corresponding row of $Q$, and restores target coordinate scales. Single-layer Procrustes gives mean correlation $+0.193$ over all 28 pairs and $+0.175$ over the 10 negative pairs, while 3-layer pooled Procrustes gives $+0.217$ and $+0.209$. To separate coordinate mixing from single-coordinate correspondence, we also evaluate the support-matched top-$k$ truncation frontier (Table~\ref{tab:natural-signs} and Section~\ref{sec:natural}) by truncating each single-layer Procrustes row $Q_{i,:}$ to its largest $k$ entries, renormalized to a matched $8\sigma$ intervention norm. Across all 28 stable pairs, mean correlation is $+0.163$ at $k=1$ (versus $+0.179$ for one-coordinate $\Bd$), $+0.167$ at $k=2$, $+0.180$ at $k=4$ (the smallest support size $k$ matching one-coordinate $\Bd$), $+0.188$ at $k=8$, $+0.200$ at $k=64$, and $+0.193$ at full $k=896$; on the 10 negative pairs, $k=1$ selects the exact same partner and negative sign as $\Bd$ on 10/10 pairs, matching its $+0.185$ correlation.

Table~\ref{tab:natural-signs} summarizes the held-out causal intervention correlations on negatively oriented, prompt-stable pairs across Qwen2/Qwen2.5-0.5B, TinyLlama old-3T/v1.1, and Qwen2/Qwen2.5-1.5B. All 10 negative Qwen2/2.5-0.5B pairs beat sign removal at every dose ($2\sigma,4\sigma,8\sigma$; 1,000-draw prompt-block bootstrap 95\% CI $[+0.185,+0.203]$ at $8\sigma$), and all 3 negative TinyLlama pairs do so at both tested doses ($2\sigma,8\sigma$).

\begin{table}[ht]
\centering\small
\setlength{\tabcolsep}{3.5pt}
\caption{Held-out causal intervention correlations on negatively oriented, prompt-stable coordinate pairs across distinct pretraining runs (higher is better). Signed ($\Bd$) applies the negative fitted sign; Sign rem.\ forces $+1$ on the same partner; Fitted $\Sd$ uses a separate unsigned assignment; Random perturbs a random coordinate (or shuffled-pair control on Qwen2/2.5-1.5B, where random coordinate is $-0.011$); Dense $O(d)$ applies a full orthogonal Procrustes row.}
\label{tab:natural-signs}
\begin{tabular}{llrrrrrr}
\toprule
Model pair (neg/stable/$d$) & Dose & Signed $\Bd$ & Sign rem. & Fitted $\Sd$ & Random & Dense $O(d)$ & Wins\\
\midrule
Qwen2/2.5-0.5B (10/28/896) & $2\sigma$ & $+0.132$ & $-0.112$ & -- & -- & -- & 10/10\\
& $4\sigma$ & $+0.154$ & $-0.083$ & -- & -- & -- & 10/10\\
& $8\sigma$ & $+0.185$ & $-0.010$ & $+0.047$ & $+0.022$ & $+0.209$ & 10/10\\
\midrule
TinyLlama 3T/v1.1 (3/10/2048) & $2\sigma$ & $+0.118$ & $-0.106$ & $+0.061$ & $-0.016$ & $+0.120$ & 3/3\\
& $8\sigma$ & $+0.169$ & $-0.051$ & $+0.097$ & $+0.002$ & $+0.133$ & 3/3\\
\midrule
Qwen2/2.5-1.5B (6/15/1536) & $8\sigma$ & $+0.189$ & $-0.021$ & -- & $+0.055$ & -- & 6/6\\
\bottomrule
\end{tabular}
\end{table}

The TinyLlama replication compares \texttt{TinyLlama-1.1B-intermediate-step-1431k-3T} and \texttt{TinyLlama\_v1.1}, matching block outputs 5, 11, and 16 and intervening at block output 11. It retains 10 prompt-stable pairs (3 negative). Fitting the permutation-only ($\Sd$) map across pooled depths gives negative-subset correlations $+0.069$ and $+0.095$ at $2\sigma$ and $8\sigma$ ($+0.061$ and $+0.097$ single-layer), while pooled dense $O(d)$ gives $+0.120$ and $+0.133$ ($+0.100$ and $+0.108$ single-layer).

At 1.5B, four-bank agreement selects 15 of 1,536 Qwen2/Qwen2.5 pairs (6 negative), while selecting solely on the three non-evaluation banks W1/M/C (holding out W2 strictly for evaluation) returns 17 pairs that contain all 15. On held-out W2 at $8\sigma$, signed $\Bd$ transfer beats sign removal on $15/15$ pairs overall (mean $+0.140$ vs.\ $-0.004$; shuffled-pair control $+0.052$, winning $13/15$) and on $6/6$ negative-sign pairs (mean $+0.189$ vs.\ $-0.021$; shuffled-pair control $+0.055$, random coordinate $-0.011$).

\subsection{Global rejection and sign-rate control}

For the global Qwen1.5B test, 70,656 token rows from W1 are pooled over layers 7, 14, and 21 to fit one full-space map, and 70,656 held-out W2 rows provide evaluation (standardized by W1 means and scales). We compute
\[
R^2=1-\frac{\|B_2-A_2M\|_F^2}{\|B_2\|_F^2},
\]
where the $\Sd$, $\Bd$, and $O(d)$ maps optimize their respective W1 objectives and ridge regularization ($\lambda=10^4$) is selected on a split of W1 before refitting on all of W1. Negative full-space discrete $R^2$ ($-0.971$ Identity, $-0.742$ $\Sd$, $-0.722$ $\Bd$) shows that cross-generation correspondences are strictly sparse rather than global.

In the Qwen2.5 Base/Math prompt-stability check, W1 selects 29 negative pairs, of which 13 recur on W2, 0 on M, and 1 on C with the same partner and sign (0/29 across all four banks). Separately, permuting token rows to destroy pairing while preserving per-model coordinate statistics yields a fitted negative-sign fraction of $0.497\pm 0.008$ over 12 refits, confirming that a $\sim 50\%$ raw negative-sign rate is the pairing-destroyed null rather than evidence of natural sign flips.

\subsection{Natural merging and a planted-coordinate-change control}
\label{app:natural-merge}

The Qwen2.5-0.5B natural merge panel fine-tunes separate descendants on AG News (250 steps) and Yelp (150 steps) with effective batch size 32 and learning rate $10^{-5}$. Table~\ref{tab:natural-merge-panel} compares their unmodified 50/50 weight average in inherited slots against task arithmetic ($\lambda\in\{0.25,0.5,0.75,1.0\}$) and TIES ($t\in\{0.2,0.5,0.8\}$), followed by a controlled test in which an exact $\Bd$ relabeling is planted on one descendant prior to merging.

\begin{table}[ht]
\centering\small
\caption{Natural descendant merging and planted-relabeling control on Qwen2.5-0.5B (classification accuracy: higher is better; WikiText NLL: lower is better). Inherited average (50/50 weight average) and Task arithmetic with $\lambda=0.5$ are algebraically identical ($\theta_0+0.5(\theta_A-\theta_0)+0.5(\theta_B-\theta_0)=0.5(\theta_A+\theta_B)$). We show the full tested merge grid rather than a held-out hyperparameter split; in the planted-relabeling control, Fitted $\Sd$ estimates an unsigned permutation from activations (which alters partner selection as well as signs, unlike a known-permutation sign ablation).}
\label{tab:natural-merge-panel}
\begin{tabular}{lrrr}
\toprule
Method & AG News acc. & Yelp acc. & WikiText NLL\\
\midrule
Inherited average & 0.9119 & 0.5133 & 2.877\\
Task arithmetic, $\lambda=0.25$ & 0.8969 & 0.5600 & 2.835\\
Task arithmetic, $\lambda=0.5$ & 0.9119 & 0.5133 & 2.877\\
Task arithmetic, $\lambda=0.75$ & 0.9131 & 0.4800 & 2.954\\
Task arithmetic, $\lambda=1$ & 0.9113 & 0.4633 & 3.076\\
TIES, $t=0.2$ & 0.9163 & 0.4933 & 2.932\\
TIES, $t=0.5$ & 0.9163 & 0.4833 & 2.943\\
TIES, $t=0.8$ & 0.9181 & 0.5300 & 2.893\\
\midrule
\multicolumn{4}{l}{After a planted signed relabeling of one descendant}\\
Uncorrected average & 0.2700 & 0.1967 & 13.025\\
Fitted $\Sd$ & 0.2494 & 0.1833 & 13.070\\
Fitted $\Bd$ & 0.9125 & 0.5233 & 2.880\\
Gain-folded $O(d)$ & 0.9056 & 0.5067 & 2.909\\
\bottomrule
\end{tabular}
\end{table}

\FloatBarrier
\section{Trajectory budgets, checkpoint spacing, and retained labels}
\label{app:trajectory-resources}

\begin{table}[ht]
\centering\small
\caption{Mean coordinate recovery (\%, higher is better) over the dependent nine-pair Qwen2.5-0.5B panel under matched activation-row budgets $B$ (where $B$ caps total activation-row evaluations rather than distinct prompts or runtime). All six checkpoint strides are reported because selecting their maximum is a post-hoc oracle choice; the best path is ahead at two budgets, tied at one, and behind at one, so these near-ceiling means support no robust ordering.}
\label{tab:budget-panel}
\begin{tabular}{rrrrrrrr}
\toprule
$B$ & Endpoint & Gap 10 & 20 & 40 & 50 & 100 & 200\\
\midrule
1,000 & 100.000 & 99.926 & 100.000 & 100.000 & 98.500 & 96.193 & 98.562\\
2,000 & 99.975 & 100.000 & 99.380 & 98.264 & 98.946 & 98.053 & 98.512\\
4,000 & 99.926 & 99.814 & 96.491 & 99.206 & 98.996 & 99.653 & 99.306\\
7,901 & 99.901 & 99.690 & 97.445 & 100.000 & 99.926 & 99.926 & 96.553\\
\bottomrule
\end{tabular}
\end{table}

\begin{table}[ht]
\centering\small
\caption{Checkpoint-stride sweep on a smooth 200-step Qwen2.5-0.5B run (\%, higher is better), measuring the percentage of coordinates returned to their inherited index across a fixed prompt set and training trajectory with no injected relabeling.}
\label{tab:spacing}
\begin{tabular}{rrr}
\toprule
Checkpoint gap (steps) & Fits & Recovery (\%)\\
\midrule
2 & 100 & 100.00\\
4 & 50 & 99.22\\
10 & 20 & 100.00\\
20 & 10 & 98.10\\
50 & 4 & 99.33\\
100 & 2 & 99.00\\
200 & 1 & 99.22\\
\bottomrule
\end{tabular}
\end{table}

\begin{table}[ht]
\centering\small
\caption{Cross-run coordinate recovery (\%, higher is better) across 18 dependent pairs from nine shared runs per horizon, comparing single-shot endpoint matching and 2-step path composition (which requires intermediate checkpoints and activation fits) against the exact carry-forward baseline (Retained initial). Endpoint wins counts pairs with strictly higher endpoint accuracy than path accuracy.}
\label{tab:carry-forward}
\begin{tabular}{lrrrrr}
\toprule
Model & Steps & Retained initial & Endpoint & Path & Endpoint wins\\
\midrule
Llama-3.2-1B & 1,500 & 100.0 & 70.3 & 66.9 & 9/18\\
Llama-3.2-1B & 200 & 100.0 & 70.8 & 88.6 & 0/18\\
Qwen2.5-1.5B & 1,500 & 100.0 & 60.3 & 91.1 & 6/18\\
Qwen2.5-1.5B & 200 & 100.0 & 61.3 & 96.2 & 6/18\\
\bottomrule
\end{tabular}
\end{table}

\paragraph{Row-budget panel (Table~\ref{tab:budget-panel}).}
The nine-pair Qwen2.5-0.5B panel crosses six 200-step fine-tuning runs (so the nine pairs are dependent, and we report means without confidence intervals). Each run trains in \texttt{float32} on a V100-SXM2-32GB GPU at layer 12 with batch size 4, sequence length 128, learning rate $5\times10^{-5}$ with 10 warmup steps, masked padding labels, data shards interleaved independently of the relabeling arm, and a recorded step-0 gauge (seed 17). Correlation matching uses activation captures spaced by 10 steps and subsampled to evaluate strides of 10, 20, 40, 50, 100, and 200 steps; we report all six strides because selecting the best spacing after observing the result is an oracle comparison rather than a deployable selection rule.

\paragraph{Smooth-run spacing sweep (Table~\ref{tab:spacing}).}
The separate 200-step Qwen2.5-0.5B spacing sweep (layer 12, batch size 4, learning rate $2\times10^{-4}$, fixed 6,144-token probe set) measures return to inherited indices with no explicit relabeling during training. Because local errors can either cancel or compound across steps, end-to-end recovery does not follow a universal product-of-edge-accuracies law, and this smooth trajectory does not test checkpoint spacing around the abrupt mid-trajectory fit failures seen in the 1,500-step Llama-3.2-1B run.

\paragraph{Resource and carry-forward accounting (Table~\ref{tab:carry-forward}).}
In our cost profile across two 200-step runs on an RTX 3090 Ti (captures every 2 steps at one layer on one prompt set), each activation capture occupies 143,249,024 bytes; retaining all 202 captures on disk therefore uses 28,936,302,848 bytes (28.9\,GB), whereas streaming two captures uses 286,498,048 bytes (286.5\,MB). A 201-fit cross-run chain (100 transitions per run plus one base link) and a single endpoint fit are timed on already captured activations, excluding model loading, forward passes, and extraction; consequently, matching activation-row counts $B$ does not equalize end-to-end compute or storage cost.

\FloatBarrier
\section{Exact gain-tie census}
\label{app:gain-census}

We inspect the stored input, post-attention, and final RMSNorm gain tensors of five public checkpoints. Equality is exact in the stored numeric values, with no rounding or tolerance. For each residual index we concatenate its gains across all these normalizations, then count identical tuples. Each inspected model has zero repeated joint tuples across all $2L+1$ normalizations, even though individual normalization layers contain tied entries.

\begin{table}[ht]
\centering\small
\caption{Exact stored-gain census across five public RMSNorm checkpoints. Largest single tie counts coordinates sharing a single value within one normalization layer, whereas Joint tied pairs counts coordinate pairs equal across every normalization layer; observing zero joint ties verifies joint distinctness (Theorem~\ref{thm:native-gauge}) for each inspected revision despite layerwise ties.}
\label{tab:gain-census}
\begin{tabular}{lrrrr}
\toprule
Checkpoint & Width & Norms & Largest single tie & Joint tied pairs\\
\midrule
Llama-3.2-1B & 2048 & 33 & 345 & 0\\
Llama-3.2-1B-Instruct & 2048 & 33 & 308 & 0\\
SmolLM2-360M & 960 & 65 & 124 & 0\\
SmolLM2-360M-Instruct & 960 & 65 & 108 & 0\\
TinyLlama-1.1B old-3T & 2048 & 45 & 2 & 0\\
\bottomrule
\end{tabular}
\end{table}

For reproducibility, our release includes the checkpoint revision identifiers, a SHA-256 digest of every inspected normalization tensor, and the census script, which reads the relevant \texttt{safetensors} payloads directly and verifies all $2L+1$ residual normalization gains on saved model parameters without training or forward passes.

\FloatBarrier
\section{Detailed limitations and broader impacts}
\label{app:limitations}

\paragraph{Detailed limitations.}
First, our exact theoretical results characterize normalization-covariant orthogonal reparameterizations and the induced transport rules for coordinate-indexed artifacts, rather than unconstrained input--output network identifiability. Second, natural-checkpoint alignment depends on checkpoint lineage: inherited coordinate slots provide a strong default across same-base continued-pretraining and instruction-tuned descendants, whereas cross-generation pairs support only sparse, prompt-stable coordinate subsets rather than global discrete permutations, and dense $O(d)$ rotations can achieve higher overall reconstruction when coordinate mixing is permitted. Third, our trajectory experiments evaluate step-wise local $\Bd$ composition against recorded step-0 relabelings (where carrying forward the known initial map is 100\% exact and local transition fits require intermediate checkpoints and probe evaluations), while Proposition~\ref{prop:no-canonical} shows that no canonical coordinate transport exists between independently specified parameterizations from function-level information alone. Fourth, while our stored-gain census verifies that all five inspected checkpoints have zero joint gain ties across layers, near-ties can reduce finite-sample activation assignment margins.

\paragraph{Broader impacts.}
Gauge-correct transport can improve reproducibility and checkpoint-editing audits, but can also lower the cost of transferring behavioral interventions, including undesirable steering directions, between related checkpoints. We therefore present it as an audit and verification tool, and emphasize gauge-invariant reporting when coordinate identity is unnecessary.

\end{document}